\documentclass[10pt,twocolumn,letterpaper]{article}

\usepackage[pagenumbers]{cvpr} 

\definecolor{cvprblue}{rgb}{0.21,0.49,0.74}
\usepackage[pagebackref,breaklinks,colorlinks,allcolors=cvprblue]{hyperref}
\usepackage{diagbox}
\usepackage{multirow}
\usepackage{appendix}
\usepackage{graphicx}
\usepackage{booktabs}

\usepackage{xcolor}
\usepackage{amsmath}
\definecolor{customgreen}{rgb}{0.0, 0.5, 0.0}
\definecolor{customred}{rgb}{0.6, 0.0, 0.0}

\def\paperID{17562} 
\def\confName{CVPR}
\def\confYear{2025}

\title{Thinking Before Looking: \\Improving Multimodal LLM Reasoning via Mitigating Visual Hallucination}

\author{
    \vspace{-2mm}
    Haojie Zheng$^{1}$\footnotemark[1] \footnotemark[2]\quad 
    Tianyang Xu$^{2}$ \footnotemark[1] \footnotemark[2] \quad
    Hanchi Sun$^{3}$\quad
    Shu Pu$^{4}$\quad
    Ruoxi Chen$^{3}$\footnotemark[2]\quad 
    Lichao Sun$^{3}$ \\
    \vspace{-2mm}
    $^{1}$University of Pennsylvania\quad 
    $^{2}$Columbia University\quad \\
    \vspace{-2mm}
    $^{3}$Lehigh University\quad 
    $^{4}$Independent Researcher \\
    \vspace{-2mm}
    {\tt\small haojiez@seas.upenn.edu, tx2240@columbia.edu, pushuabc@gmail.com}\\
    {\tt\small{has423,lis221}@lehigh.edu,chenrx0830@gmail.com}
}

\begin{document}
\maketitle
\footnotetext[1]{These authors contributed equally to this work.}
\footnotetext[2]{Visiting student at Lehigh University.}
\begin{abstract}
Multimodal large language models (MLLMs) have advanced the integration of visual and linguistic modalities, establishing themselves as the dominant paradigm for visual-language tasks. Current approaches like chain of thought (CoT) reasoning have augmented the cognitive capabilities of large language models (LLMs), yet their adaptation to MLLMs is hindered by heightened risks of hallucination in cross-modality comprehension. In this paper, we find that the \textit{thinking while looking} paradigm in current multimodal CoT approaches—where reasoning chains are generated alongside visual input—fails to mitigate hallucinations caused by misleading images. To address these limitations, we propose the \textbf{Visual Inference Chain (VIC)} framework, a novel approach that constructs reasoning chains using textual context alone before introducing visual input, effectively reducing cross-modal biases and enhancing multimodal reasoning accuracy. Comprehensive evaluations demonstrate that VIC significantly improves zero-shot performance across various vision-related tasks, mitigating hallucinations while refining the reasoning capabilities of MLLMs. Our anonymized code repository can be found at \url{https://github.com/Terry-Xu-666/visual_inference_chain}.
\end{abstract}

\section{Introduction}
\label{introdution}

\begin{figure*}[ht]
    \centering
    \includegraphics[width=\textwidth]{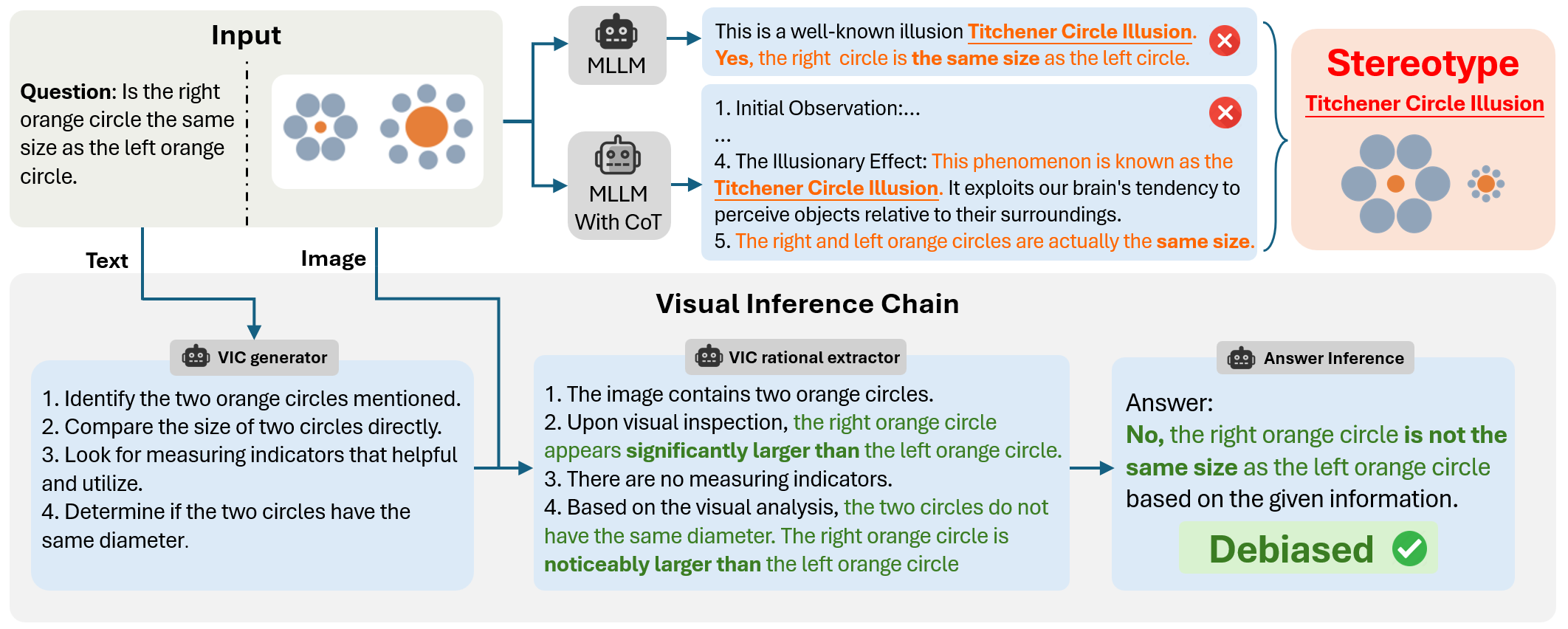}
    \caption{This example from HallusionBench demonstrates the differences between zero-shot, zero-shot CoT, and VIC. The zero-shot CoT represents the \textit{thinking while looking} approach, which tends to exhibit stereotypical reasoning patterns when processing both visual and textual inputs simultaneously. In contrast, our \textit{thinking before looking} paradigm, VIC, enhances reasoning quality by decoupling the visual and textual inputs. More examples can be found in Appendix E.}
    \label{fig:1}
\end{figure*}

Large Language Models (LLMs), such as GPT-4 \cite{achiam2023gpt} and Llama \cite{touvron2023llama}, have driven remarkable advancements in world comprehension~\cite{hao2023reasoninglanguagemodelplanning} and reasoning capability~\cite{huang-chang-2023-towards}. 
Prompting techniques like CoT~\cite{wei2022chain,rubin2021learning,fu2022complexity} have been developed to enhance LLMs' ability to handle complex tasks through human-like step-by-step reasoning.
Meanwhile, MLLMs have rapidly advanced in recent years~\cite{achiam2023gpt,bai2023qwen,liu2024improved,liu2024visual}, extending LLMs’ capabilities into the multimodal realm by integrating visual backbones to align visual and language representations.
MLLMs have demonstrated exceptional performance in a range of vision-related tasks, including visual question answering~\cite{bansal2020visual}, object recognition~\cite{du2022learning,zang2024contextual}, and video comprehension~\cite{liu2024tempcompass}, highlighting the impressive evolution of AI-driven visual-language understanding. The success of CoT prompting in unimodal contexts suggests a promising extension to multimodal scenarios. Due to the integration of pretrained vision models \cite{radford2021learning,sun2023eva} and language models \cite{zhang2023llama} in MLLMs, various types of hallucinations have been observed \cite{li2023evaluating,liu2023aligning,cadene2019rubi}, including nonexistent object generation \cite{tong2024eyes}, visual misinterpretations \cite{liu2023hallusionbench}, and cross-modality biases \cite{vosoughi2024cross}. 

Offering advantages in tackling complex multimodal tasks, current prompting approaches \cite{zhang2023multimodal, zheng2023ddcot} predominantly adhere to the \textit{thinking while looking} paradigm, where reasoning occurs simultaneously with the integration of visual elements. However, this paradigm encounters substantial challenges due to the prevalence of hallucinations, which undermine both the reliability of the reasoning process and the accuracy of the responses. As illustrated in Figure \ref{fig:1}, the MLLM often falls into stereotypes when processing a question-image pair, where it recalls similar prior contexts and overlooks subtle variations, leading to erroneous responses. Even with CoT prompting, the model tends to rely on memory-based stereotypes rather than engaging in accurate reasoning, which leads to incorrect responses.

To overcome hallucinations from visual inputs and harness the reasoning capabilities of LLMs, we propose the Visual Inference Chain (VIC), which introduces a reasoning process that occurs prior to engaging with visual elements, following the \textit{thinking before looking} paradigm. This approach mirrors human cognition \citep{solso2005cognitive, summerfield2009expectation}, where reasoning often precedes perception. For example, when adults hear a question before seeing the image, they generate a preliminary plan based on their accumulated experience. The question activates relevant memories and contextual knowledge, allowing them to deduce a forward-looking reasoning strategy ahead of engaging in visual stereotypes. This underscores that the direct impact of visual elements on the reasoning process is relatively limited. Analogous to human cognition, MLLMs can adopt the \textit{thinking before looking} paradigm. By leveraging this paradigm, VIC taps into the accumulated forward-looking reasoning capabilities of powerful LLMs, enabling MLLMs to anticipate and recognize patterns more efficiently by dynamically adjusting reasoning steps. Additionally, the VIC framework employs a systematic multi-step detachment strategy to minimize compounding hallucinations from both the question and the image, further enhancing the MLLMs' reasoning capabilities for complex tasks.

In this work, we demonstrate that our \textit{thinking before looking} strategy outperforms the conventional \textit{thinking while looking} approach in multimodal reasoning tasks. Our method demonstrates improved performance in empirical studies on GPT-series~\cite{openai2024gpt4o} and Gemini-series~\cite{deepmind2024gemini} models across various visual question benchmarks. VIC achieves notable improvements on hallucination-specific benchmarks such as MMVP~\citep{tong2024eyes}, HallusionBench~\citep{guan2024hallusionbench}, and POPE~\citep{li2023evaluating}, as well as on general multimodal benchmarks including MME~\citep{fu2024mmecomprehensiveevaluationbenchmark}, MathVista~\citep{lu2023mathvista}, and SEED-Bench~\citep{li2024seed}. For instance, the Gemini 1.5 Pro sees a substantial improvement of $\boldsymbol{31.74\%}$ on the MMVP benchmark, in a meanwhile the GPT-4o mini model shows an increase of $\boldsymbol{16.59\%}$. Across all benchmarks, GPT-series models average a $\boldsymbol{8.02\%}$ refinement and Gemini-series models show an average improvement of $\boldsymbol{7.19\%}$, underscoring both the effectiveness and robustness of our method.
\section{Related Works}
\label{relatedWork}

\paragraph{CoT reasoning with LLMs.} Chain-of-thought (CoT) reasoning \citep{wei2022chain} has significantly enhanced large language models (LLMs) performance by guiding them to break down complex tasks into intermediate reasoning steps. Subsequent work introduced self-consistency \citep{wang2022self} where multiple reasoning paths are generated, and the most consistent answer is chosen. Recent advances focus on optimizing CoT with example selection and enhanced reasoning frameworks \citep{zhang2022automatic}, exploring selecting diverse examples to guide CoT reasoning. Step-Aware Verifier framework enhances CoT by incorporating a verification step at each intermediate reasoning stage, improving reasoning reliability \citep{li2022making}. Meanwhile, Tree of Thought (ToT) \citep{yao2024tree} extends CoT by exploring multiple reasoning paths in a tree structure, ensuring thorough consideration of potential solutions. Other refinements allow models to self-reflect on and adjust their reasoning steps \citep{xie2024self}.

\paragraph{Multimodal CoT reasoning.} Considering the natural gap between text and vision data, Multimodal Large Language Models (MLLMs) are less capable of utilizing the reasoning ability of CoT. MMCoT \citep{zhang2023multimodal} adds a rationale generation block before answer inference. DDCoT \citep{zheng2023ddcot} is proposed to enhance multimodal reasoning by assigning specific tasks to language and visual models. Meanwhile, Visual CoT \citep{shao2024visual} incorporates the notion of segmentation, and performs well for object-detection tasks but lacks versatility. Image-of-Thought \citep{zhou2024image} further explores the MLLMs CoT prompting by introducing specific actions like segmentation, zoom-in, and color-space conversions in the CoT chain, extending the applicable scenarios with the sacrifice of flexibility. Compositional CoT \citep{mitra2024compositional} extracts the compositional information of images as a scene graph (SG) and then utilize it during the CoT process. Other works like CoCoT~\citep{zhang2024cocot} and KAM-CoT~\citep{mondal2024kam}, consider multiple input image scenarios and incorporate knowledge graphs (KG) of multimodal information. However, many such techniques rely on fixed templates for extracting predetermined information, resulting in a lack of flexibility and the utilization of reasoning capabilities. The straightforward \textit{thinking while looking} approach struggles to resolve problematic cross-modal interactions, often resulting in hallucinations.

\paragraph{Hallucination in MLLMs.} Hallucination remains a significant challenge in MLLMs, where models generate information that is factually incorrect or irrelevant to the given inputs \citep{bai2024hallucination}. In multimodal reasoning, hallucinations can occur when models produce textual outputs not grounded in the visual data, leading to fabricated details in tasks such as image captioning or visual question answering (VQA).
For hallucination evaluation, FaithScore \citep{jing2023faithscore} extracts fine-grained atomic facts from the generated answer and then conducts consistency verifications. POPE, a polling-based query method \citep{li2023evaluating} is proposed for a better evaluation of object hallucination. Recently, HallusionBench \citep{guan2024hallusionbench} further explored the hallucination evaluation, by editing the original input image and forming different text-image pairs to diagnose failure types from language hallucination and visual illusion. Several hallucination-mitigating methods are also proposed. To conduct instruction tuning that mitigate hallucination, LRV \citep{liu2023mitigating} serves as the first large and diverse visual instruction tuning dataset and VIGC \citep{wang2024vigc} aims for data generation. As for training-free method, Woodpecker \citep{yin2023woodpecker} uses fixed steps to correct hallucination in MLLMs. Our method, however, generates applicable reasoning trajectories without visual elements for a given question, which is more flexible and accurate than fixed analyzing steps, and could be further utilized across different VQA tasks.
\section{Method}

\begin{figure*}[ht]
    \centering
    \includegraphics[width=\textwidth]{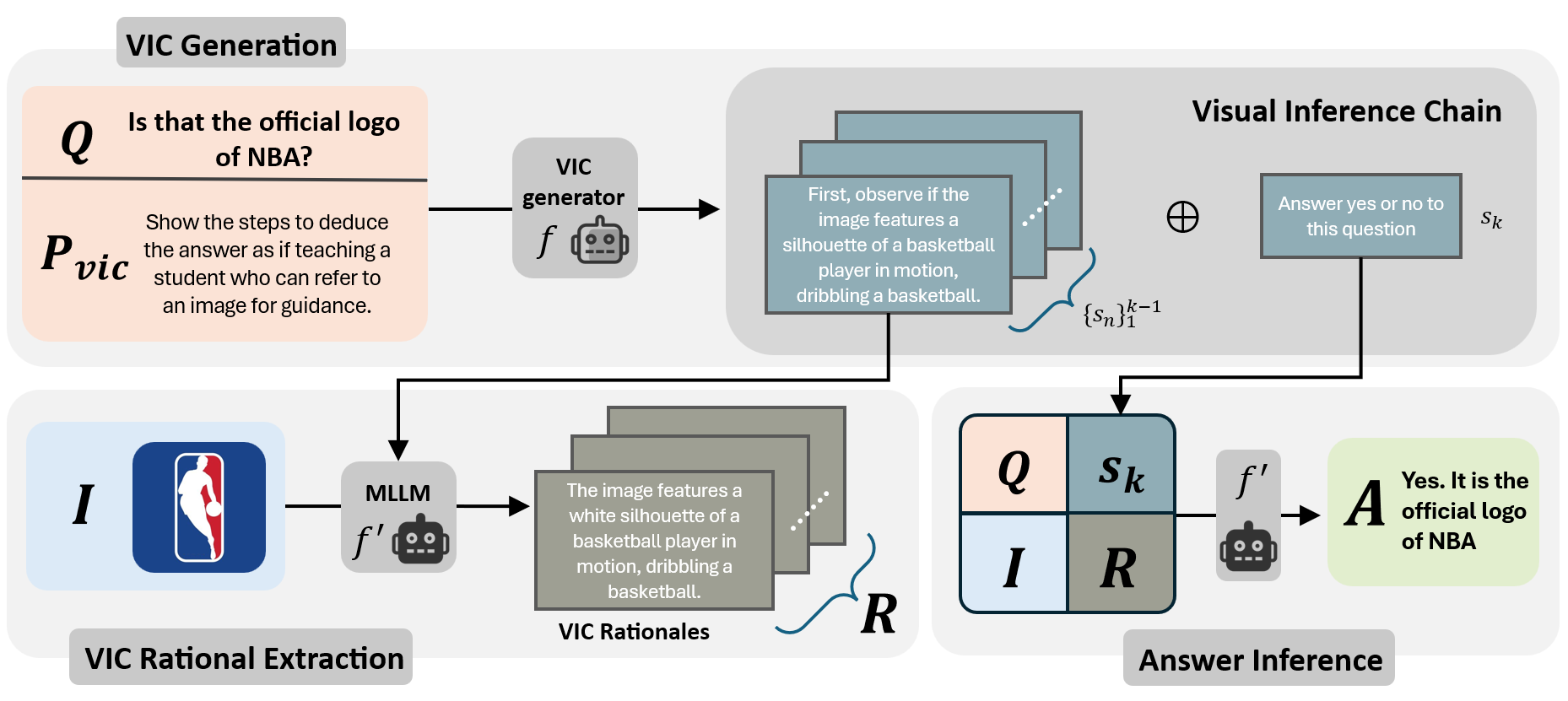}
    \caption{\textbf{The overall framework of VIC (Visual Inference Chain)}. VIC decouples visual and textual inputs to improve reasoning. It first generates intermediate reasoning steps from the question \(\mathbf{Q}\) and prompt \(\mathbf{P}_{vic}\). The image \(\mathbf{I}\) is processed through an MLLM to extract rationales \(\mathbf{R}\), which, combined with the visual inference steps \(\{s_n\}_{1}^{k}\), lead to the final answer \(\mathbf{A}\) with enhanced accuracy.}
    \label{fig:2}
\end{figure*}

In this section, we begin by introducing the preliminary architecture of MLLMs and the process of \textit{thinking while looking} paradigm in Section \ref{MLLM} and Section \ref{Thinking while looking} respectively. In Section~\ref{VIC}, we present the foundation of \textit{thinking before looking} paradigm along with a detailed explanation of how forward-looking reasoning can enhance the reasoning process more effectively than \textit{thinking while looking} paradigm. Additionally, we explore two key components in the framework
in Section \ref{information} and \ref{answer}, respectively. The detailed implementation of the prompting mechanism discussed in this section is provided in Appendix D, and the overview of our method is illustrated in Figure~\ref{fig:2}.


\subsection{Multimodal LLM}
\label{MLLM}

MLLMs are designed to handle data from various modalities, such as text, images, and audio, enabling the integration and generation of multimodal information. In the task of Visual Question Answering (VQA), the MLLM is provided with an image input \(\mathbf{I}\) and a textual input \(\mathbf{Q}\). These inputs are then encoded into a shared representation space using a pre-trained visual encoding model, and a fixed text tokenization process for the textual data. The encoded representations from both visual and textual data are subsequently processed by large language model. We characterize the whole MLLM as \(f'(*)\), which has been trained on unified representation data from both modalities. During inference, the model generates a response \(\mathbf{A}\) given \(\mathbf{I}\) and \(\mathbf{Q}\), which can be formalized as:

\begin{align}
\label{eq:1}
\mathbf{A} &= f'(\mathbf{I}, \mathbf{Q}).
\end{align}

For the sake of simplicity, we ignore the tokenizer and pre-trained visual encoding model since they primarily serve as preprocessing components. This formulation allows MLLMs to focus on leveraging the combined information from multiple data types, enabling them to effectively process and respond to complex, multimodal queries with greater efficiency.

\subsection{Thinking while looking}
\label{Thinking while looking}
A simple VQA inference process is illustrated in Equation~\ref{eq:1}. Moreover, the \textit{thinking while looking} paradigm involves step-by-step reasoning trajectory while simultaneously processing the visual input. This approach generates a reasoning strategy chain \(\{s_n\}_{n=1}^{k}\), accompanied by corresponding rationales, denoted as \(\{r_n\}_{n=1}^{k}\), where the value of \(k\) varies dynamically depending on both the specific model and the given question-image pair. In parallel with generating these reasoning steps and rationales, the model also produces the final answer \(A\).

\begin{align}
\label{eq:twl}
(\{s_n,r_n\}_{n=1}^{k},\mathbf{A}) &= f'(\mathbf{I}, \mathbf{Q},\mathbf{P}_{cot}).
\end{align}

In this formula, \(\{s_n, r_n\}_{n=1}^{k}\) represents the sequence of reasoning steps and corresponding rationales. \(\mathbf{P}_{cot}\) denotes the prompt used for CoT prompting, which can be either a zero-shot prompt such as ``Let’s think step by step" or a few-shot prompt designed in an in-context learning manner. 

Although Equation \ref{eq:twl} demonstrates how CoT elucidates the MLLM's thinking process and enhances its capacity to manage complex tasks, the tight interweaving of textual and visual inputs introduces hallucination issues, as the reasoning sequence \(\{s_n, r_n\}_{n=1}^{k}\) can become biased. As shown in Figure \ref{fig:1}, the model generates a hallucination by identifying a non-existent Tichener circle illusion due to the combined influence of visual cues and the question, which activates memories of the most similar context from its training phase. This example highlights the biased reasoning steps produced by the vanilla CoT method, leading to improper reasoning in the model's responses.

\subsection{VIC generation}
\label{VIC}
The limitations of the \textit{thinking while looking} paradigm inspire the development of the \textit{thinking before looking} paradigm. This new approach promotes forward-looking reasoning, enhancing the reasoning process and improving the quality of rationality. Separating visual and textual inputs allows for more structured thinking and clearer cognitive steps. This approach decouples the question from the image, reducing bias in the reasoning steps sequence and enhancing overall reasoning performance.



The language models have internalized extensive reasoning knowledge from large-scale pre-trained data, allowing them to gain beneficial insights into image reasoning analysis. Due to their highly developed pattern recognition abilities, these models can accumulate sufficient background knowledge and generate reasonable reasoning steps for visual elements. Thus, LLMs \(f(*)\) can automatically generate forward-looking reasoning steps as an average over a broader context of similar situations, triggered by the input \((\mathbf{Q}, \mathbf{P}_{vic})\), rather than focusing solely on a specific input pair \((\mathbf{Q}, \mathbf{I})\). This process discretizes the reasoning steps from the specific input pair, reducing hallucinations while maintaining the validity of instructions for image analysis, as these instructions align with the aggregate of the most relevant contextual knowledge. Consequently, the visual inference process can be expressed as follows.

\begin{align}
    \{s_n\}_{n=1}^{k} &= f(\mathbf{Q},\mathbf{P}_{vic})
\end{align}

By employing this specific prompt, we generate a visual inference chain, denoted as \(\{s_n\}_{n=1}^{k}\), which serves as the reasoning process derived exclusively from the given question. This process can be divided into two main components. The initial \(k-1\) steps, \(\{s_n\}_{n=1}^{k-1}\), correspond to the sequence of instructions related to both recognition and reasoning. And the final step, \(s_k\), of the visual inference chain introduces a format instruction based on the specific question. This step is critical in further enhancing the framework's ability to follow complex instructions effectively.

The advantage of the \textit{thinking before looking} phase is that it eliminates the immediate need for visual information. As a result, the model \(f(*)\) can function as either a large language model or a multimodal language model operating in a blind mode. This concept of \textit{thinking before looking} allows us to leverage the superior reasoning capabilities of advanced large language models, offering significant flexibility and generalizability.

\subsection{VIC rationale extraction}
\label{information}
Hallucinations commonly arise from deep entanglement between image and textual inputs. In this step, we decouple the original question and requiring the MLLM to recognize and follow the visual inference chain step by step, mitigating the effects of textual bias. Moreover, the visual inference chain provides a more precise trajectory compared to the original question, necessitating detailed extraction of relevant information and thus reducing the risk of hallucinations by visual inference chain. 

Unlike the previous step, which can use any modality model \(f\), this step specifically employs the MLLM \(f'\). Leveraging the multitask-following capability of closed-form models, we generate the VIC rationales in a single step to produce the entire set of rationales \(\{r_n\}_{n=1}^{k-1}\), denoted collectively as \(\mathbf{R}\). We further explore the differences between single-step rationale extraction and multi-step rationale extraction in our ablation experiments.

\begin{align}
    \mathbf{R} &= f'(\mathbf{I},\mathbf{P}_{extract}, \{s_n\}_{n=1}^{k-1})
\end{align}

In this formula, \(\mathbf{I}\) represents the image corresponding to the specific question \(\mathbf{Q}\). The sequence \(\{s_n\}_{n=1}^{k-1}\) refers to the first \(k-1\) steps of the process, excluding the format instructions. \(\mathbf{P}_{extract}\) is a prompt designed to integrate the image with the visual inference chain, facilitating effective information extraction.

\subsection{Answer Inference}
\label{answer}

In the final step, we provide the same inputs to the MLLM \(f_{\phi}\) as used during the VIC rationale generation phase. These inputs include the original question-image pair \(\mathbf{Q}, \mathbf{I}\), the VIC rationale results \(\mathbf{R}\), the format instruction \(s_k\), and the reflection prompt \(\mathbf{P}_{reflect}\).

\begin{align}
    \mathbf{A} &= f'(\mathbf{I}, \mathbf{Q} , \mathbf{R} ,\mathbf{P}_{reflect}, s_k)
\end{align}

This process incorporates the VIC rationale results as additional information to support the model's response. Furthermore, we introduce a reflection mechanism at this stage, encouraging the model to reconsider the question, image, and VIC rationale results, rather than treating the rationale as the definitive answer. The format instruction \(s_k\) ensures the response adheres to the desired format by guiding the model’s analysis of the user's question or query, thereby improving instruction-following performance. 
\section{Experiment}

\begin{table*}[ht]
    \centering
    \resizebox{\textwidth}{!}{%
    \begin{tabular}{cccccccccc}
    \toprule
        \multirow{2}{*}{\textbf{Models}} & \multirow{2}{*}{\textbf{Method}} & \multirow{2}{*}{\textbf{MMVP}} & \multirow{2}{*}{\textbf{HallusionBench}} & \multirow{2}{*}{\textbf{POPE\textsubscript{adversarial}}} & \multirow{2}{*}{\textbf{Mathvista}} & \multirow{2}{*}{\textbf{SEED-Bench\textsubscript{single}}} & \multirow{2}{*}{\textbf{Average}} & \multicolumn{2}{c}{\textbf{MME}} \\
        \cmidrule(lr){9-10}
        & & & & & & & & \textbf{Perception} & \textbf{Cognition} \\
    \midrule
        \multirow{2}{*}{\textit{Baselines}} & Human & 0.957 & 0.986 & 0.995 & 0.603 & 0.967 & 0.901 & \textbf{-} & \textbf{-} \\
        & Random choice & 0.250 & 0.500 & 0.500 & 0.179 & 0.250 & 0.336 & \textbf{-} & \textbf{-} \\
    \midrule
        \multirow{3}{*}{GPT-4o mini} & Origin & 0.446 & 0.574 & 0.786 & 0.526 & 0.636 & 0.590 & 1097.23 & 407.14 \\
        & zero-shot CoT & 0.443 \textsubscript{\textcolor{customred}{($\downarrow$ 2.85\%)}} & 0.611 \textsubscript{\textcolor{customgreen}{($\uparrow$ 6.46\%)}} & 0.773 \textsubscript{\textcolor{customred}{($\downarrow$ 1.65\%)}} & 0.520 \textsubscript{\textcolor{customred}{($\downarrow$ 1.14\%)}} & 0.660 \textsubscript{\textcolor{customgreen}{($\uparrow$ 3.77\%)}} & 0.601 \textsubscript{\textcolor{customgreen}{($\uparrow$ 1.86\%)}} & 1069.81 \textsubscript{\textcolor{customred}{($\downarrow$ 2.50\%)}} & 417.14 \textsubscript{\textcolor{customgreen}{($\uparrow$ 2.46\%)}} \\
        & VIC & \textbf{0.520} \textsubscript{\textcolor{customgreen}{($\uparrow$ 16.59\%)}} & \textbf{0.639} \textsubscript{\textcolor{customgreen}{($\uparrow$ 11.37\%)}} & \textbf{0.793} \textsubscript{\textcolor{customgreen}{($\uparrow$ 0.89\%)}} & \textbf{0.536} \textsubscript{\textcolor{customgreen}{($\uparrow$ 1.90\%)}} & \textbf{0.696} \textsubscript{\textcolor{customgreen}{($\uparrow$ 9.43\%)}} & \textbf{0.637} \textsubscript{\textcolor{customgreen}{($\uparrow$ 7.96\%)}} & \textbf{1105.27} \textsubscript{\textcolor{customgreen}{($\uparrow$ 0.73\%)}} & \textbf{505.00} \textsubscript{\textcolor{customgreen}{($\uparrow$ 24.04\%)}} \\
    \midrule
        \multirow{3}{*}{Gemini 1.5 Flash} & Origin & 0.527 & 0.560 & 0.769 & 0.479 & 0.658 & 0.599 & 1077.36 & 358.92 \\
        & zero-shot CoT & 0.513 \textsubscript{\textcolor{customred}{($\downarrow$ 2.54\%)}} & 0.614 \textsubscript{\textcolor{customgreen}{($\uparrow$ 9.69\%)}} & 0.741 \textsubscript{\textcolor{customred}{($\downarrow$ 3.64\%)}} & \textbf{0.520} \textsubscript{\textcolor{customgreen}{($\uparrow$ 8.56\%)}} & 0.672 \textsubscript{\textcolor{customgreen}{($\uparrow$ 2.13\%)}} & 0.612 \textsubscript{\textcolor{customgreen}{($\uparrow$ 2.17\%)}} & 1105.9 \textsubscript{\textcolor{customgreen}{($\uparrow$ 2.65\%)}} & 500.71 \textsubscript{\textcolor{customgreen}{($\uparrow$ 39.50\%)}} \\
        & VIC & \textbf{0.553} \textsubscript{\textcolor{customgreen}{($\uparrow$ 5.05\%)}} & \textbf{0.638} \textsubscript{\textcolor{customgreen}{($\uparrow$ 13.93\%)}} & \textbf{0.780} \textsubscript{\textcolor{customgreen}{($\uparrow$ 1.43\%)}} & 0.516 \textsubscript{\textcolor{customgreen}{($\uparrow$ 7.72\%)}} & \textbf{0.713} \textsubscript{\textcolor{customgreen}{($\uparrow$ 8.36\%)}} & \textbf{0.640} \textsubscript{\textcolor{customgreen}{($\uparrow$ 6.84\%)}} & \textbf{1118.54} \textsubscript{\textcolor{customgreen}{($\uparrow$ 3.82\%)}} & \textbf{508.21} \textsubscript{\textcolor{customgreen}{($\uparrow$ 41.59\%)}} \\
    \midrule
        \multirow{3}{*}{GPT-4o} & Origin & 0.673 & 0.626 & 0.811 & 0.597 & 0.657 & 0.673 & 1174.39 & 522.85 \\
        & zero-shot CoT & 0.687 \textsubscript{\textcolor{customgreen}{($\uparrow$ 1.99\%)}} & 0.673 \textsubscript{\textcolor{customgreen}{($\uparrow$ 7.49\%)}} & 0.793 \textsubscript{\textcolor{customred}{($\downarrow$ 2.22\%)}} & \textbf{0.622} \textsubscript{\textcolor{customgreen}{($\uparrow$ 4.19\%)}} & 0.739 \textsubscript{\textcolor{customgreen}{($\uparrow$ 12.48\%)}} & 0.701 \textsubscript{\textcolor{customgreen}{($\uparrow$ 4.16\%)}} & 1166.12 \textsubscript{\textcolor{customred}{($\downarrow$ 0.70\%)}} & 537.14 \textsubscript{\textcolor{customgreen}{($\uparrow$ 2.73\%)}} \\
        & VIC & \textbf{0.747} \textsubscript{\textcolor{customgreen}{($\uparrow$ 10.90\%)}} & \textbf{0.692} \textsubscript{\textcolor{customgreen}{($\uparrow$ 10.52\%)}} & \textbf{0.827} \textsubscript{\textcolor{customgreen}{($\uparrow$ 1.27\%)}} & 0.620 \textsubscript{\textcolor{customgreen}{($\uparrow$ 3.85\%)}} & \textbf{0.751} \textsubscript{\textcolor{customgreen}{($\uparrow$ 14.31\%)}} & \textbf{0.727} \textsubscript{\textcolor{customgreen}{($\uparrow$ 8.08\%)}} & \textbf{1238.69} \textsubscript{\textcolor{customgreen}{($\uparrow$ 5.48\%)}} & \textbf{557.85} \textsubscript{\textcolor{customgreen}{($\uparrow$ 6.69\%)}} \\
    \midrule
        \multirow{3}{*}{Gemini 1.5 Pro} & Origin & 0.420 & 0.617 & 0.779 & \textbf{0.568} & 0.678 & 0.612 & \textbf{1166.82} & 462.14 \\
        & zero-shot CoT & 0.500 \textsubscript{\textcolor{customgreen}{($\uparrow$ 19.05\%)}} & 0.611 \textsubscript{\textcolor{customred}{($\downarrow$ 0.97\%)}} & 0.793 \textsubscript{\textcolor{customgreen}{($\uparrow$ 1.67\%)}} & 0.563 \textsubscript{\textcolor{customred}{($\downarrow$ 0.88\%)}} & 0.691 \textsubscript{\textcolor{customgreen}{($\uparrow$ 1.92\%)}} & 0.632 \textsubscript{\textcolor{customgreen}{($\uparrow$ 3.20\%)}} & 1099.33 \textsubscript{\textcolor{customred}{($\downarrow$ 5.78\%)}} & 521.42 \textsubscript{\textcolor{customgreen}{($\uparrow$ 12.83\%)}} \\
        & VIC & \textbf{0.553} \textsubscript{\textcolor{customgreen}{($\uparrow$ 31.74\%)}} & \textbf{0.664} \textsubscript{\textcolor{customgreen}{($\uparrow$ 7.62\%)}} & \textbf{0.803} \textsubscript{\textcolor{customgreen}{($\uparrow$ 3.08\%)}} & 0.558 \textsubscript{\textcolor{customred}{($\downarrow$ 1.76\%)}} & \textbf{0.713} \textsubscript{\textcolor{customgreen}{($\uparrow$ 5.16\%)}} & \textbf{0.658} \textsubscript{\textcolor{customgreen}{($\uparrow$ 7.55\%)}} & 1147.23 \textsubscript{\textcolor{customred}{($\downarrow$ 1.68\%)}} & \textbf{561.42} \textsubscript{\textcolor{customgreen}{($\uparrow$ 21.48\%)}} \\
    \bottomrule
    \end{tabular}%
    }
    \caption{\textbf{Main Results Across Models and Benchmarks.} Comparison of model performance on benchmarks such as MMVP, HallusionBench, and Mathvista, including metrics for Perception and Cognition under MME. Improvements are marked in green and declines in red. Bold values denote the best results.}
    \label{table:main_results}
\end{table*}

\subsection{Experiment Setup}
\paragraph{Datasets} 
Our framework is evaluated on six benchmark datasets. To evaluate the generality and versatility of the VIC framework, we tested it across two key benchmark categories: (1) Hallucination detection benchmark including \textbf{HallusionBench} \citep{guan2024hallusionbench}, \textbf{MMVP} \citep{tong2024eyes}, and \textbf{POPE} \citep{li2023evaluating}, all of which are designed to analyze hallucinations that are prone to occur in different forms. (2) General multimodal capability benchmark, \textbf{MME} \citep{fu2024mmecomprehensiveevaluationbenchmark} and \textbf{SEED-Bench} \citep{li2024seed}, which cover general and comprehensive forms of visual question answering. \textbf{MathVista} \citep{lu2023mathvista} that focuses on math element recognition and reasoning problems, evaluates the compound capability to solve visual math challenges. We discuss the details and implementations for each benchmark in Appendix A.
\vspace{-2mm}
\paragraph{Baseline}
In our experiments, we compare the proposed VIC methodology with two primary baselines, as outlined in Table \ref{table:main_results}. The first baseline involves applying the model directly to each benchmark without incorporating any specialized techniques, enabling us to evaluate the added benefit of our method to the pretrained model's performance. The second baseline is zero-shot CoT prompt engineering. In this approach, we append the reasoning prompt such as "Let's think step by step" to the input, following the question. Since the model's output in this case includes multiple reasoning steps, we then pass it through an answer extractor to derive the final answer, in a manner similar to the final stage of the VIC method. This comparison highlights the differences and advantages of our \textit{thinking before looking} strategy. Additionally, we introduce human performance and random choices as further baselines, providing a basic performance reference point for each benchmark.
\vspace{-2mm}
\paragraph{Implementation} 
We conducted our experiments primarily on four popular closed-source MLLMs: Gemini 1.5 Flash, Gemini 1.5 Pro, GPT-4o, and GPT-4o mini. The reason for choosing these models mainly involved two aspects. (i) Our approach is grounded in the belief that the \textit{thinking before looking} capability should not be exclusive to large models, driving us to experiment on both large and small models. (ii) Our method also leverages the long instructions following ability, which is a key strength of closed-source models. The experiments were conducted in three steps. Firstly, we utilized the VIC-prompt, where the input question text was fed into the models to generate Visual Inference Chain. These generated instructions were then paired with the original image to extract VIC rationals. Finally, we derive the answer to the question based on the rationale we extracted.
Additionally, we also ran several complementary experiments on open-source models, such as Qwen-VL. For further details and results from these experiments, please refer to Section \ref{open Sorce model}.

\subsection{Results}

The results of the Visual-Inference-Chain on six different benchmarks using two open-source models are listed in Table \ref{table:main_results}.
We primarily use average accuracy as the evaluation metric to quantify the models' performance, except for the MME benchmark, where we retain the original composite evaluation score that combines accuracy and a refined accuracy metric (Accuracy+). This approach allows for a consistent comparison with previous records on MME. For the MMVP benchmark, we use pair accuracy to measure the model's ability to overcome the hallucination caused by CLIP.

\begin{figure}[ht]
  \includegraphics[width=\columnwidth]{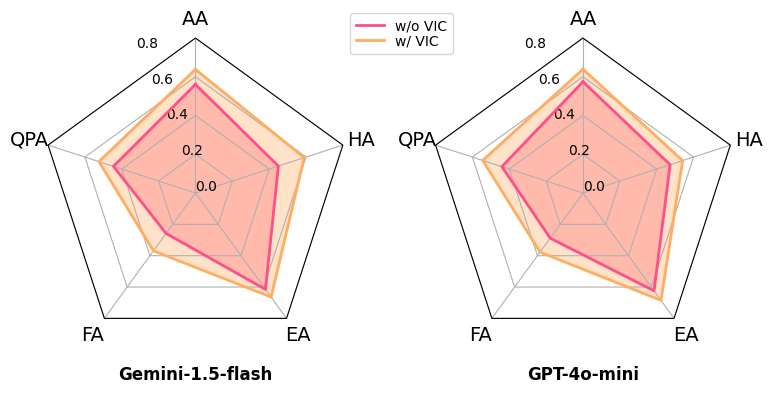}
  \caption{Detailed evaluation comparisons with and without VIC for two models on HallusionBench.  Features: AA - All Accuracy, HA - Hard Accuracy, EA - Easy Accuracy, FA - Figure Accuracy, QPA - Question Pair Accuracy.}
  \label{fig:experiments}
\end{figure}
Across the comprehensive benchmarks MME and SEED-Bench, the models show an average performance improvement of $\boldsymbol{9.13\%}$, demonstrating that our method VIC consistently enhances performance across all tasks. These results highlight the efficacy of VIC in boosting multimodal reasoning capabilities. For the Mathvista benchmark, although the average gain is slightly reduced to $\boldsymbol{3.15\%}$, this is largely due to the nature of this benchmark, which lacks detailed explanations of questions and is heavily reliant on visual information. For example, questions like "\textit{Is R0 greater than 0?}" require direct visual interpretation, making it difficult to generate an effective visual inference chain without explicit vision inputs. Despite these challenges, the improvement underscores the versatility of VIC, even in tasks that are highly vision-dependent.
\begin{figure}[ht]
  \includegraphics[width=\columnwidth]{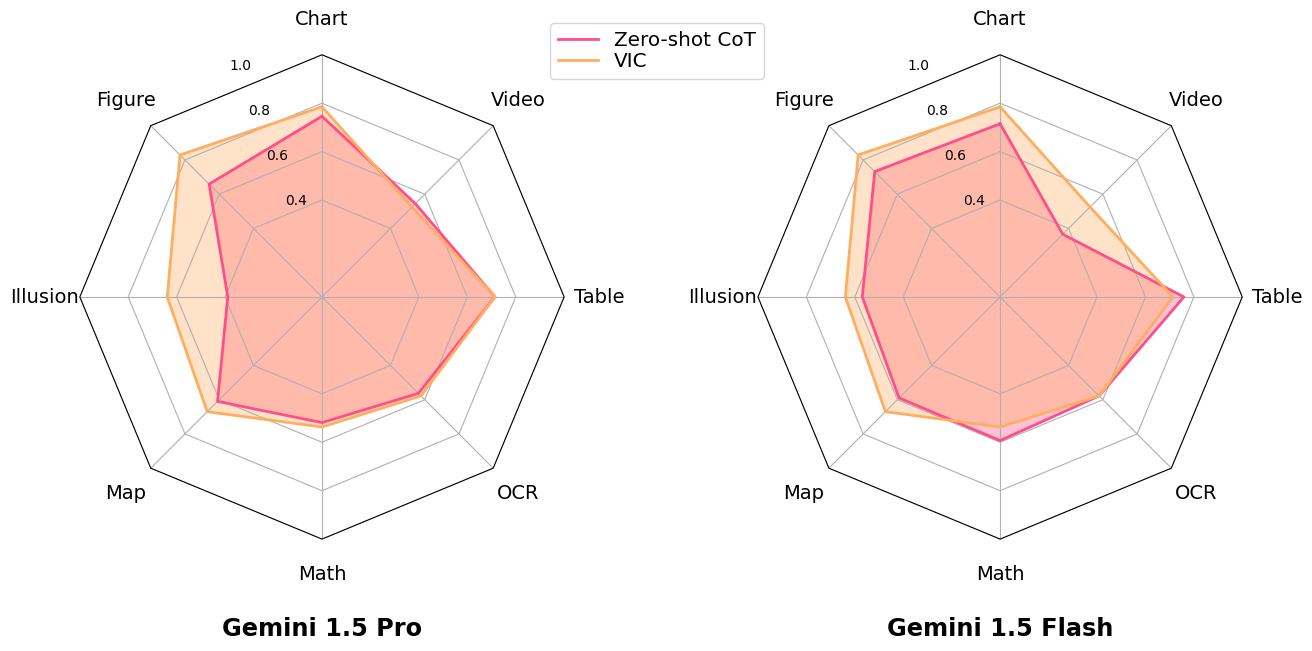}
  \caption{Performance comparison of Gemini 1.5 Flash and Gemini 1.5 Pro using Zero-shot CoT and VIC methods across various evaluation metrics on  HallusionBench. }
  \label{fig:experiments2}
\end{figure}
Regarding the hallucination benchmarks, we first test on adversarial samples in POPE. Our method still effectively handles these challenging scenarios, improving the models' performance by $\boldsymbol{1.36\%}$. As for MMVP, the average improvement achieves $\boldsymbol{15.62\%}$, which marks that VIC significantly overcomes the hallucination caused by using CLIP as a vision encoder. HallusionBench is another critical benchmark we want to focus on, the details of HallusionBench experiments are shown in Figure \ref{fig:experiments}. where we observe notable improvements in both Question Pair Accuracy and Figure Accuracy, indicating that VIC helps mitigate both language hallucinations and visual illusions. Another key observation is the enhancement in Hard Accuracy, which measures the models' ability to interpret human-edited images from HallusionBench. 

The categorization results of Gemini 1.5 Flash and Gemini 1.5 Pro on HallusionBench are presented in Figure \ref{fig:experiments2}. Compared to Zero-shot CoT, VIC demonstrates significant improvements in figure understanding, reduction of illusion errors, and enhanced map identification, while maintaining strong performance across other tasks. This result highlights VIC’s ability to significantly improve the robustness of the models, positioning it as a vital innovation in anti-hallucination methods. More details on the experiment can be found in Appendix B.

\begin{figure}[ht]
    \centering
    \includegraphics[width=\columnwidth]{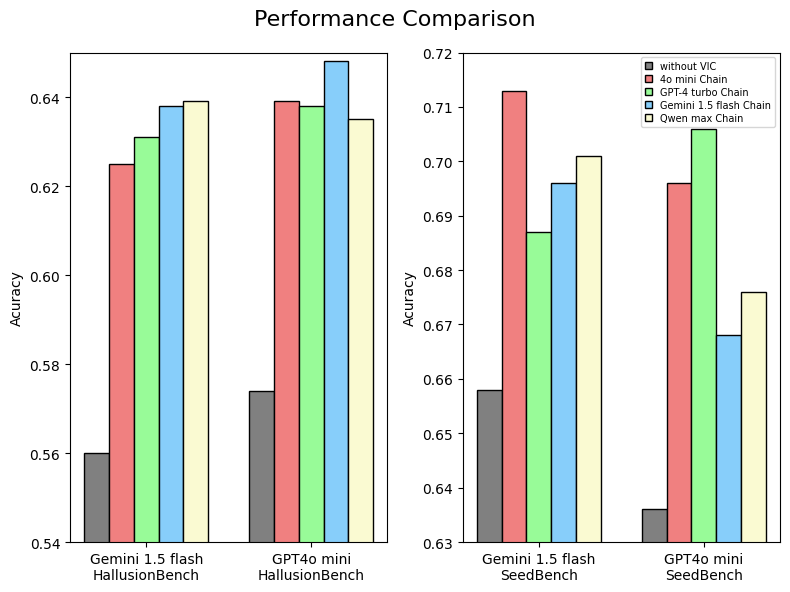}
    \caption{\textbf{Performance of Different VIC Generators.} This chart compares the performance of various VIC generators on HallusionBench and SEED-Bench. Grey bars represent the original performance for reference.}
    \label{fig:3}
\end{figure}

\subsection{Ablation Experiment}
By adopting a blind input approach, where the model operates independently of the image, we can leverage various models for VIC generation. In our experiments, we employed a range of models, including pure language models such as GPT-4 Turbo and Qwen-Max, alongside multimodal models like GPT-4o mini and Gemini 1.5 Flash, as visual inference chain generators. Additionally, we conducted ablation experiments to investigate the differences between single-step and multi-step VIC rationale extraction, the impact of reflective prompting during the answer generation phase, and the performance of open-source models. For more detailed results, please refer to Appendix C.

\paragraph{Different VIC generator.}
As shown in Figure \ref{fig:3}, different visual inference chain (VIC) generators significantly affect VIC performance. Leveraging the "thinking before looking" paradigm, we can select either a pure language model or a multimodal model operating in a blind mode. Thus, we tested GPT-4 Turbo, Qwen-Max, GPT-4o mini, and Gemini 1.5 Flash as VIC generators on two benchmarks, HallusionBench and SEED-Bench. On SEED-Bench, the maximum performance variation is 3.95\% for Gemini 1.5 Flash and 5.98\% for GPT-4o mini. A key finding is the minimal performance difference between chains generated by pure language models and those by multimodal models. For instance, Qwen-Max’s visual inference chain performs best for Gemini 1.5 Flash on HallusionBench, while GPT-4o mini’s chain achieves the highest score for Gemini 1.5 Flash on SEED-Bench. This suggests that both pure language and multimodal models offer comparable reasoning capabilities for image-related tasks, even though multimodal models are more specialized in image comprehension. Overall, there is no clear pattern in which chain consistently performs best, as effectiveness varies based on factors such as the compatibility of the visual inference chain with the MLLM and the VIC generator's ability to address specific question types. For example, the GPT-4o mini visual chain excels on general questions in SEED-Bench but performs the weakest on HallusionBench.

\paragraph{One step VIC rationale extraction.} The VIC generation process uses two main methods: a single-step approach and a multi-step approach. In the single-step method, the entire inference chain is processed as a unified input, while the multi-step approach completes tasks incrementally, integrating intermediate results at each step. Evaluations on the HallusionBench and SEED-Bench benchmarks show distinct advantages for each approach. Overall, the single-step method yields more stable improvements, ensuring greater consistency across tasks. This approach also reduces latency and computational cost due to its single-input nature. In contrast, the multi-step approach requires re-inputting images and prompts at each step, which increases response times and resource demands. Although the multi-step method has some advantages on benchmarks like HallusionBench, where detailed, stepwise reasoning can improve accuracy, the single-step approach remains generally superior in terms of efficiency, accuracy, and reliability, especially on models like Gemini 1.5 Flash and GPT-4o mini, achieving notable performance gains on SEED-Bench without the risk of error propagation.

\begin{table}[ht]
\centering
\resizebox{\columnwidth}{!}{%
\begin{tabular}{llcc}
\toprule
Model & Methods & Hallusionbench & SEED-bench \\
\midrule
Gemini 1.5 Flash & Original & 0.560 & 0.658 \\
 & Vic M & \textbf{0.66} \textsubscript{\textcolor{customgreen}{($\uparrow$ 17.86\%)}} & 0.676 \textsubscript{\textcolor{customgreen}{($\uparrow$ 2.74\%)}} \\
 & Vic One Step & 0.638 \textsubscript{\textcolor{customgreen}{($\uparrow$ 13.93\%)}} & \textbf{0.701} \textsubscript{\textcolor{customgreen}{($\uparrow$ 6.53\%)}} \\
\midrule
GPT-4o Mini & Original & 0.574 & 0.636 \\
 & Vic M & \textbf{0.673} \textsubscript{\textcolor{customgreen}{($\uparrow$ 17.21\%)}} & 0.629\textsubscript{\textcolor{customred}{($\downarrow$ 1.10\%)}} \\
 & Vic One Step & 0.648 \textsubscript{\textcolor{customgreen}{($\uparrow$ 12.92\%)}} &\textbf{ 0.676} \textsubscript{\textcolor{customgreen}{($\uparrow$ 6.29\%)}} \\
\bottomrule
\end{tabular}%
}
\caption{Performance comparison across models and methods on Hallusionbench and SEED-bench.}
\end{table}
\vspace{-6mm}
\paragraph{Reflection prompt.}
In the final stage of our approach, a critical technique is to treat the VIC rationale not as definitive information but as an input that invites reflection within the MLLM to generate the final answer. This reflective approach encourages the model to re-evaluate the correctness of the VIC rationale, thus enhancing the robustness of the overall framework. The ablation experiment on Gemini 1.5 Flash in SEED-Bench shows the impact of this reflection prompt: removing it leads to a slight performance decrease of 0.3\%, yet the model still achieves a notable 6.23\% improvement over the baseline without the VIC framework. As shown in Table \ref{tab:gemini-performance-transposed}, the VIC framework significantly surpasses the baseline across various tasks, particularly in areas like accuracy, spatial reasoning, and instance identity. For example, accuracy rises from 0.658 with the Origin model to 0.699 with VIC (without reflection) and further to 0.701 with full reflection. The VIC framework also enhances spatial reasoning, with scores improving from 0.419 in the baseline to 0.605 in the full setup. The inclusion of the reflection prompt solidifies the model’s robustness, as it promotes careful reassessment of initial answers, contributing to more accurate and dependable predictions across tasks.

\begin{table}[ht]
    \centering
    \resizebox{\columnwidth}{!}{%
    \begin{tabular}{lcccc}
        \toprule
        \multirow{2}{*}{\diagbox{\textit{Category}}{\textit{Method}}}& \multicolumn{4}{c}{\textbf{Gemini 1.5 Flash}} \\
        \cmidrule{2-5}
         ~& Origin & COT\textsubscript{vanilla} & VIC w/o reflection prompt & VIC \\ \midrule
        Acc & 0.658 & 0.672 & 0.699 & \textbf{0.701} \\
        Instance Attributes & 0.677 & 0.692 & 0.714 & \textbf{0.711} \\
        Instance Identity & 0.763 & 0.734 & 0.777 & \textbf{0.806} \\
        Instance Interaction & 0.900 & 0.700 & 0.800 & 0.800 \\
        Instance Location & 0.500 & 0.484 & 0.578 & \textbf{0.594} \\
        Instances Counting & 0.542 & 0.603 & \textbf{0.648} & 0.626 \\
        Scene Understanding & 0.724 & 0.724 & 0.710 & 0.719 \\
        Spatial Relation & 0.419 & 0.558 & 0.581 & \textbf{0.605} \\
        Text Understanding & 0.500 & 0.500 & 1.000 & 1.000 \\
        Visual Reasoning & \textbf{0.821} & 0.786 & 0.786 & 0.750 \\
        \bottomrule
    \end{tabular}%
    }
    \caption{Performance metrics for different methods of Gemini 1.5 Flash on Seed-Bench}
    \label{tab:gemini-performance-transposed}
\end{table}
\vspace{-3mm}
\paragraph{Open source model.}
\label{open Sorce model}
We tested the VIC framework on an open-source model, Qwen-VL Plus, an MLLM with a medium parameter size. For some benchmarks, the performance with the VIC framework declined, while others still showed significant improvement. We believe this is because the VIC framework requires a high capacity in areas such as multi-step instruction following and the ability to handle long contexts with visual input. The lack of these capabilities can likely lead to the failure of the VIC framework. Therefore, we prioritized using closed-source models with stronger overall performance across a wider range of capabilities to implement our framework effectively.
\section{Discussion}
Although our \textit{thinking before looking} framework demonstrates remarkable performance across diverse vision-text tasks, we view it as a crucial enhancement rather than the ultimate solution for MLLMs reasoning. It provides a fresh perspective in this domain and holds potential for integration with the \textit{thinking while looking} paradigm to achieve superior outcomes. In comprehensive evaluations of these two approaches, we found that the performance of \textit{thinking before looking} lagged behind \textit{thinking while looking} in certain categories. We suggest that future advancements should prioritize merging these paradigms through a selection mechanism or a process of mutual reflection. Such a combination would allow the paradigms to complement each other effectively, addressing their individual limitations and paving the way for more sophisticated reasoning strategies.

\section{Conclusion}

In this paper, we introduce the \textbf{Visual Inference Chain (VIC)} framework, a novel method that mitigates reasoning biases in MLLMs by decoupling visual and textual inputs, advancing the \textit{thinking before looking} paradigm. By systematically separating visual elements, VIC framework enhances reasoning robustness, significantly reduces hallucinations, and improves performance across diverse vision-language tasks. Our experiments demonstrate that VIC consistently boosts zero-shot performance and provides detailed insights into the impact of different VIC generators, underscoring the effectiveness of reflective prompting. Overall, VIC framework enhances accuracy and strengthens the reliability of multimodal reasoning effectively.



\newpage
{
    \small
    \bibliographystyle{ieeenat_fullname}
    \bibliography{main}
}
\appendix
\section{Dataset details}
\label{Dataset details}

\paragraph{MMVP.} MMVP is designed to evaluate MLLMs' visual capability, the benchmark is organized into nine different visual patterns and there are 15 pairs of zero-shot questions. It focuses on visual errors commonly arising from models using CLIP-based vision encoders. MMVP tests the models' capability to correctly identify and interpret subtle visual features such as object relationships, orientations, and fine-grained details. By highlighting these shortcomings, the benchmark aims to provide a more rigorous evaluation framework, helping researchers identify areas where visual integration in MLLMs can be improved. We used 150 question pairs (300 questions in total) in the MMVP test set to evaluate whether our prompting method can mitigate hallucinations arising from the vision encoder.

\paragraph{HallusionBench.} HallusionBench is a diagnostic suite designed to analyze the dual issues of language hallucination and visual illusion in MLLMs. It includes 1129 handcrafted VQA pairs featuring 165 original images and 181 images expertly modified by human professionals. Since our framework is designed for single image and question pairs, we selected a subset from HallucinationBench that contains only single-image questions, excluding text-only questions and two-image questions. There are 674 single image questions in our selected subset in total.

\paragraph{MME.} MME is a comprehensive benchmark that measures both perception and cognition across 14 subtasks. All tasks involve "Yes-or-No" questions, and the scoring combines accuracy and a refined accuracy metric (accuracy+). The total scores for perception and cognition are 2000 and 800, respectively, reflecting a broad assessment of MLLM performance in both areas.  We used all the questions in MME to test our framework.

\paragraph{Mathvista.} Mathvista is introduced to systematically evaluate mathematical reasoning ability in visual contexts. It is derived from 31 different datasets and completing these tasks requires fine-grained, deep visual understanding and compositional reasoning. We used the testMini set from Mathvista, which contains 1,000 visual math questions.

\paragraph{POPE.} POPE focuses on assessing object hallucinations in MLLMs. The dataset includes three sampling strategies: random, popular, and adversarial. Since the random and popular strategies are less challenging, our experiment concentrates on 1000 VQA pairs generated from the adversarial strategy, which presents more difficult test cases. 

\paragraph{SEED-Bench.} SEED-Bench serves as a comprehensive benchmark to assess the generative capabilities of multimodal large language models. It focuses on evaluating models through tasks requiring deep comprehension across both visual and textual inputs. SEED-Bench provides a challenging environment by introducing novel generative tasks, like free-form answer generation, image understanding, and reasoning. To evaluate our framework on SEED-Bench, we randomly sampled 1,000 single-image instances from the SEED-Bench dataset as our test set.

To maintain consistency and accuracy in the evaluation, we developed a unified evaluation framework by reconstructing the previous six benchmark evaluation metrics. This framework is designed to handle multiple-choice questions (MCQ), yes/no (YORN) questions, and mixed-form questions. Our evaluation framework uses a large language model to extract answers, parse options, yes/no signals, and more. This approach has been widely adopted in various benchmark evaluations, as models like GPT-3.5 and GPT-4o mini can achieve nearly 100\% accuracy in such answer-parsing tasks, as demonstrated in several previous works. Therefore, we employ this evaluation framework for the VIC framework, and the code for our evaluation is available in our repository.

\section{Detailed Results and Analysis}
\subsection{Hallusionbench}

\begin{table*}[ht!]
\centering
    \resizebox{\textwidth}{!}{%
    \begin{tabular}{ccccccccccccc}
    \toprule
    \textbf{model} & \textbf{method} & \textbf{Acc} & \textbf{chart} & \textbf{figure} & \textbf{illusion} & \textbf{map} & \textbf{math} & \textbf{ocr} & \textbf{table} & \textbf{video} & \textbf{precision} & \textbf{Yes\_rate} \\
    \midrule
    GPT-4o mini & origin & 0.574 & 0.585 & 0.585 & 0.625 & 0.516 & 0.426 & 0.620 & 0.741 & 0.406 & 0.543  & 0.481 \\
    ~ & zero-shot CoT & 0.611 & 0.638 & \textbf{0.610} & 0.611 & \textbf{0.578} & 0.519 & 0.630 & \textbf{0.848} & 0.366 & \textbf{0.595} &  0.454 \\
    ~ & VIC & \textbf{0.639} & \textbf{0.723} & 0.512 & \textbf{0.694} & 0.547 & \textbf{0.648} & \textbf{0.660} & 0.777 & \textbf{0.426} & 0.576  & 0.438 \\
    \midrule
    Gemini 1.5 Flash & origin & 0.560 & 0.623 & 0.732 & 0.597 & 0.516 & \textbf{0.648} & 0.450 & 0.661 & 0.366 & 0.559  & 0.350 \\
    ~ & zero-shot CoT & 0.614 & \textbf{0.715} & 0.732 & 0.569 & 0.578 & 0.593 & \textbf{0.590} & \textbf{0.759} & 0.366 & 0.564  & 0.429 \\
    ~ & VIC & \textbf{0.625} & 0.669 & 0.732 & \textbf{0.625} & \textbf{0.625} & 0.630 & 0.570 & 0.714 & \textbf{0.475} & \textbf{0.612}  & 0.352 \\
    \midrule
    GPT-4o  & origin & 0.626 & 0.738 & 0.780 & 0.403 & 0.594 & 0.537 & 0.610 & 0.804 & 0.465 & 0.579  & 0.433 \\
    ~ & zero-shot CoT & 0.673 & \textbf{0.777} & \textbf{0.854} & 0.542 & 0.578 & \textbf{0.574} & 0.670 & \textbf{0.848} & \textbf{0.495} & \textbf{0.634}  & 0.405 \\
    ~ & VIC & \textbf{0.692} & 0.754 & 0.756 & \textbf{0.667} & \textbf{0.625} & 0.519 & \textbf{0.800} & 0.750 & 0.455 & 0.625  & 0.408 \\
    \midrule
    Gemini 1.5 Pro & origin & 0.617 & 0.677 & 0.683 & 0.389 & \textbf{0.609} & \textbf{0.704} & 0.610 & 0.705 & \textbf{0.545} & 0.612  & 0.352 \\
    ~ & zero-shot CoT & 0.611 & 0.746 & 0.659 & 0.389 & 0.563 & 0.519 & 0.610 & \textbf{0.714} & \textbf{0.545} & 0.588 & 0.386 \\
    ~ & VIC & \textbf{0.664} & \textbf{0.785} & \textbf{0.829} & \textbf{0.639} & 0.578 & 0.537 & \textbf{0.670} & \textbf{0.714} & 0.525 & \textbf{0.646} &  0.340 \\
    \bottomrule
\end{tabular}%
}
\caption{Performance comparison of different models and methods evaluated on HallusionBench.}
\label{tab:hallucinationbench}
\end{table*}

Table \ref{tab:hallucinationbench} presents detailed experimental results of various models on HallucinationBench, evaluating performance across subtasks such as chart, figure, illusion, map, and others. Overall, the VIC method consistently outperforms both the original and zero-shot CoT methods in terms of accuracy, highlighting its effectiveness across the models. Notably, for the table and figure subtasks, zero-shot CoT slightly surpasses VIC. This can be attributed to the fact that these subtasks rely heavily on visual information, which diminishes the advantage of VIC’s \textit{Thinking before looking} paradigm. However, VIC demonstrates superior performance in most other tasks, with its average results surpassing zero-shot CoT. Particularly in the illusion subtask, VIC's dominance underscores its strength in mitigating hallucinations, making it highly effective in handling anti-hallucination tasks.

\subsection{POPE adversarial}

\begin{table}[ht!]
\centering
    \resizebox{\columnwidth}{!}{%
    \begin{tabular}{ccccccc}
    \toprule
    \textbf{model} & \textbf{method} & \textbf{Acc} & \textbf{F1} & \textbf{precision} & \textbf{recall} & \textbf{yes\_rate} \\
    \midrule
    GPT-4o mini & origin & 0.786 & \textbf{0.797} & 0.847 & \textbf{0.752} & 0.445 \\
    ~ & zero-shot CoT & 0.773 & 0.755 & 0.91 & 0.645 & 0.355 \\
    ~ & VIC & \textbf{0.793} & 0.757 & \textbf{0.92} & 0.643 & 0.35 \\
    \midrule
    Gemini 1.5 Flash & origin & 0.769 & 0.724 & 0.92 & 0.597 & 0.325 \\
    ~ & zero-shot CoT & 0.741 & 0.685 & \textbf{0.978} & 0.527 & 0.27 \\
    ~ & VIC & \textbf{0.78} & \textbf{0.751} & 0.89 & \textbf{0.649} & 0.365 \\
    \midrule
    GPT-4o & origin & 0.811 & \textbf{0.824} & 0.869 & \textbf{0.782} & 0.451 \\
    ~ & zero-shot CoT & 0.793 & 0.804 & \textbf{0.897} & 0.729 & 0.407 \\
    ~ & VIC & \textbf{0.827} & 0.822 & 0.869 & 0.78 & 0.45 \\
    \midrule
    Gemni 1.5 Pro & origin & 0.779 & 0.77 & 0.849 & 0.705 & 0.416 \\
    ~ & zero-shot CoT & 0.792 & \textbf{0.79} & \textbf{0.905} & 0.701 & 0.388 \\
    ~ & VIC & \textbf{0.803} & \textbf{0.79} & 0.888 & \textbf{0.713} & 0.402 \\
    \bottomrule
    \end{tabular}%
    }
    
\caption{Performance comparison of different models and methods.}
\label{tab:performance_comparison_pope}
\end{table}

Table \ref{tab:performance_comparison_pope} shows that VIC demonstrates its robustness in POPE adversarial samples. While zero-shot CoT exhibits higher precision, VIC maintains a better balance between precision and recall, leading to more reliable overall performance. For example, Gemini 1.5 Pro shows a precision of 0.888 with VIC compared to 0.905 with zero-shot CoT, but VIC achieves a better recall (0.713 vs. 0.701). The Yes rate remains relatively consistent across all methods, but VIC continues to provide more balanced results. This pattern of improvement across different models highlights VIC's general effectiveness in delivering robust, superior, and well-rounded model performance on POPE.
\subsection{MME}

\begin{table*}[ht!]
\centering
    \resizebox{\textwidth}{!}{%
    \begin{tabular}{ccccccccccccccccccccc}
    \toprule
    \textbf{Model} & \textbf{Method} & \textbf{Per} & \textbf{Cog} & \textbf{Ex} & \textbf{Cnt} & \textbf{Pos} & \textbf{Col} & \textbf{Post} & \textbf{Cel} & \textbf{Scn} & \textbf{Lmk} & \textbf{Art} & \textbf{OCR} & \textbf{CSR} & \textbf{NumC} & \textbf{TxtT} & \textbf{CodR} \\
    \midrule
    GPT-4o mini & Origin & 1097.23 & 407.14 & 136.67 & 115 & 78.33 & \textbf{135} & 134.35 & 45.88 & \textbf{122.5} & 91.5 & 88 & \textbf{150} & 117.14 & 87.5 & \textbf{110} & 92.5 \\
    ~ & zero-shot CoT & 1069.82 & 417.14 & 126.67 & 111.67 & 73.33 & \textbf{135} & \textbf{135.03} & 49.12 & 119.5 & 77.75 & 91.75 & \textbf{150} & 112.14 & 105 & 100 & 100 \\
    ~ & VIC & \textbf{1105.27} & \textbf{505.00} & \textbf{140} & \textbf{120} & \textbf{95} & 131.67 & 132.65 & \textbf{49.71} & 111 & \textbf{116.5} & \textbf{101.25} & 107.5 & \textbf{125} & \textbf{140} & 100 & \textbf{140} \\
    \midrule
    Gemini 1.5 Flash & Origin & 1077.37 & 358.93 & 130 & 93.33 & 58.33 & 123.33 & 118.71 & \textbf{124.41} & \textbf{117.75} & 95.5 & 101 & 115 & 101.43 & 90 & 90 & 77.5 \\
    ~ & zero-shot CoT & 1105.9 & 500.71 & 123.33 & 128.33 & 68.33 & 126.67 & \textbf{129.93} & 97.06 & 111.5 & 87.5 & 98.25 & \textbf{135} & \textbf{125.71} & 150 & 85 & \textbf{140} \\
    ~ & VIC & \textbf{1118.54} & \textbf{508.21} & \textbf{136.67} & \textbf{130} & \textbf{90} & \textbf{133.33} & 126.53 & 111.76 & 99 & \textbf{96.25} & \textbf{102.5} & 92.5 & 115.71 & \textbf{150} & \textbf{102.5} & \textbf{140} \\
    \midrule
    GPT-4o & Origin & 1174.39 & 522.86 & 143.33 & 128.33 & 101.67 & \textbf{143.33} & 140.14 & 10.59 & \textbf{115.75} & 136 & 105.25 & \textbf{150} & \textbf{132.86} & 120 & 140 & 130 \\
    ~ & zero-shot CoT & 1166.11 & 537.14 & 135 & \textbf{130} & 106.67 & 140 & \textbf{140.82} & 15.88 & 112.5 & 129 & 111.25 & 145 & 127.14 & 130 & 135 & \textbf{145} \\
    ~ & VIC & \textbf{1238.69} & \textbf{557.85} & \textbf{146.67} & 121.67 & \textbf{118.33} & 141.67 & 135.03 & \textbf{73.82} & 112.75 & \textbf{137.5} & \textbf{116.25} & 135 & 127.86 & \textbf{145} & \textbf{140} & \textbf{145} \\
    \midrule
    Gemini 1.5 Pro & Origin & \textbf{1166.82} & 462.14 & 133.33 & 120 & 76.67 & \textbf{135} & 131.63 & \textbf{127.94} & \textbf{113} & \textbf{103.5} & 100.75 & \textbf{125} & 117.14 & 115 & 100 & 130 \\
    ~ & zero-shot CoT & 1099.33 & 521.42 & 133.33 & \textbf{123.33} & 86.67 & 126.67 & 115.99 & 97.35 & 113.5 & 80.5 & \textbf{102} & 120 & 121.43 & 145 & 120 & 135 \\
    ~ & VIC & 1147.23 & \textbf{561.42} & \textbf{140} & 118.33 & \textbf{91.67} & \textbf{135} & \textbf{133.67} & 122.06 & 104.75 & 93 & 98.75 & 110 & \textbf{126.43} & \textbf{150} & \textbf{140} & \textbf{145} \\
    \bottomrule
    \end{tabular}%
    }
    
\caption{Performance comparison of different models and methods. Per: Perception, Cog: Cognition, Ex: Existence, Cnt: Count, Pos: Position, Col: Color, Post: Posters, Cel: Celebrity, Scn: Scene, Lmk: Landmark, Art: Artwork, OCR: Optical Character Recognition, CSR: Commonsense Reasoning, NumC: Numerical Calculation, TxtT: Text Translation, CodR: Code Reasoning.}
\label{tab:performance_comparison_MME}
\end{table*}

Table \ref{tab:performance_comparison_MME} presents detailed experimental results on the MME benchmark, highlighting that the VIC method consistently outperforms zero-shot CoT across a range of models in both cognition and perception tasks. VIC demonstrates significant advantages in most subtasks, such as existence, position, landmark recognition, artwork analysis, numerical calculation, text translation, and code reasoning. However, both VIC and zero-shot CoT perform poorly on tasks like OCR, color recognition, and poster identification, which do not require advanced reasoning. Overall, VIC enhances the model’s reasoning abilities, leading to superior performance on more complex and sophisticated problems. 

\subsection{Mathvista}

\begin{table*}[ht!]
\centering
    \resizebox{0.7\textwidth}{!}{%
    \begin{tabular}{cccccccccc}
    \toprule
    \textbf{model} & \textbf{method} & \textbf{Acc} & \textbf{LR} & \textbf{AR} & \textbf{GR} & \textbf{SR} & \textbf{ALR} & \textbf{SCR} & \textbf{NC} \\
    \midrule
    GPT-4o mini & origin & 0.526 & 0.236 & \textbf{0.581} & \textbf{0.611} & \textbf{0.637} & 0.189 & 0.467 & \textbf{0.582} \\
    ~ & zero-shot CoT & 0.520 & 0.243 & 0.467 & 0.586 & 0.588 & \textbf{0.601} & \textbf{0.574} & 0.271 \\
    ~ & VIC & \textbf{0.536} & \textbf{0.582} & 0.405 & 0.544 & 0.515 & 0.499 & 0.571 & 0.292 \\
    
    \midrule
    Gemini 1.5 Flash & origin & 0.479 & 0.523 & 0.519 & 0.324 & \textbf{0.566} & 0.419 & 0.518 & 0.340 \\
    ~ & zero-shot CoT & \textbf{0.520} & 0.537 & \textbf{0.615} & \textbf{0.641} & 0.324 & \textbf{0.445} & \textbf{0.519} & \textbf{0.340} \\
    ~ & VIC & 0.516 & \textbf{0.639} & 0.292 & 0.635 & 0.456 & 0.378 & 0.427 & 0.456 \\
    \midrule
    GPT-4o & origin & 0.597 & 0.216 & 0.535 & 0.389 & \textbf{0.704} & \textbf{0.664} & 0.630 & 0.607 \\
    ~ & zero-shot CoT & \textbf{0.622} & 0.297 & \textbf{0.661} & \textbf{0.721} & 0.569 & 0.389 & 0.676 & \textbf{0.689} \\
    ~ & VIC & 0.620 & \textbf{0.734} & 0.351 & 0.656 & 0.633 & 0.598 & 0.623 & 0.361 \\
    \midrule
    Gemini 1.5 Pro & origin & \textbf{0.568} & \textbf{0.687} & 0.615 & \textbf{0.665} & \textbf{0.456} & 0.299 & \textbf{0.648} & 0.270 \\
    ~ & zero-shot CoT & 0.563 & 0.619 & 0.577 & 0.664 & 0.162 & \textbf{0.711} & 0.250 & 0.470 \\
    ~ & VIC & 0.558 & 0.326 & \textbf{0.691} & 0.507 & 0.541 & 0.502 & 0.297 & \textbf{0.639} \\
    \bottomrule
    \end{tabular}%
    }
    
\caption{Performance comparison of different models and methods across various reasoning categories. Acc: Accuracy, LR: Logical Reasoning, AR: Arithmetic Reasoning, GR: Geometry Reasoning, SR: Statistical Reasoning, ALR: Algebraic Reasoning, SCR: Scientific Reasoning, NC: Numeric Commonsense.}
\label{tab:performance_comparison_reasoning}
\end{table*}

Table \ref{tab:performance_comparison_reasoning} presents the detailed evaluation results for the Mathvista benchmark, where the VIC method does not perform as well compared to other benchmarks. The primary reason for this underperformance is attributed to the lack of detailed textual information in the benchmark's questions. Many of the questions, such as \textit{Is $R0$ greater than $R$} or \textit{Find $x$}, lack sufficient explanations and background information, which undermines VIC’s ability to leverage its strength in reasoning through complex problems. The lack of contextual detail hinders the method’s potential to fully apply its inference capabilities. We believe that by expanding the question text and providing more detailed background descriptions, VIC could regain its superiority given by the \textit{Thinking before looking} paradigm in solving complex tasks within this benchmark.

\subsection{SEED-Bench}
\begin{table*}[ht!]
\centering
    \resizebox{0.8\textwidth}{!}{%
    \begin{tabular}{cccccccccccc}
    \toprule
    \textbf{model} & \textbf{method} & \textbf{Acc} & \textbf{IA} & \textbf{II} & \textbf{IntI} & \textbf{ILoc} & \textbf{ICount} & \textbf{SU} & \textbf{SR} & \textbf{TU} & \textbf{VR} \\
    \midrule
    Gemini 1.5 Flash & origin & 0.658 & 0.677 & 0.763 & \textbf{0.9} & 0.5 & 0.542 & \textbf{0.724} & 0.419 & \textbf{0.5} & \textbf{0.821} \\
    ~ & zero-shot CoT & 0.672 & 0.692 & 0.734 & 0.7 & 0.484 & 0.603 & \textbf{0.724} & 0.558 & \textbf{0.5} & 0.786 \\
    ~ & VIC & \textbf{0.713} & \textbf{0.726} & \textbf{0.820} & 0.7 & \textbf{0.531} & \textbf{0.665} & 0.719 & \textbf{0.698} & \textbf{0.5} & 0.75 \\ \midrule
    GPT-4o mini & origin & 0.636 & 0.649 & 0.763 & \textbf{0.8} & 0.391 & 0.497 & 0.714 & 0.512 & \textbf{0.5} & \textbf{0.857} \\
    ~ & zero-shot CoT & 0.660 & 0.665 & \textbf{0.770} & \textbf{0.8} & 0.422 & 0.553 & 0.729 & \textbf{0.581} & \textbf{0.5} & \textbf{0.857} \\
    ~ & VIC & \textbf{0.696} & \textbf{0.714} & 0.755 & 0.6 & \textbf{0.531} & \textbf{0.603} & \textbf{0.767} & \textbf{0.581} & \textbf{0.5} & \textbf{0.857} \\ \midrule
    GPT-4o & origin & 0.657 & 0.720 & 0.691 & \textbf{0.8} & 0.484 & 0.486 & 0.733 & 0.558 & \textbf{1} & 0.75 \\
    ~ & zero-shot CoT & 0.739 & \textbf{0.800} & \textbf{0.791} & 0.6 & 0.578 & 0.637 & \textbf{0.767} & 0.605 & \textbf{1} & \textbf{0.821} \\
    ~ & VIC & \textbf{0.751} & 0.791 & 0.784 & 0.6 & \textbf{0.609} & \textbf{0.676} & 0.757 & \textbf{0.814} & \textbf{1} & \textbf{0.821} \\ \midrule
    Gemini 1.5 Pro & origin & 0.678 & 0.680 & \textbf{0.784} & \textbf{0.8} & 0.5 & 0.598 & 0.733 & 0.558 & 0.5 & 0.786 \\
    ~ & zero-shot CoT & 0.691 & 0.692 & 0.712 & 0.6 & 0.5 & 0.665 & \textbf{0.776} & 0.535 & \textbf{1} & \textbf{0.786} \\
    ~ & VIC & \textbf{0.713} & \textbf{0.726} & 0.777 & \textbf{0.8} & \textbf{0.516} & \textbf{0.687} & 0.733 & \textbf{0.698} & 0.5 & 0.714 \\
    \bottomrule
    \end{tabular}%
    }
    
\caption{Performance comparison of different models and methods. Columns: Acc - Accuracy, IA - Instance Attributes, II - Instance Identity, IntI - Instance Interaction, ILoc - Instance Location, ICount - Instances Counting, SU - Scene Understanding, SR - Spatial Relation, TU - Text Understanding, VR - Visual Reasoning.}
\label{tab:performance_comparison_full}
\end{table*}

Table \ref{tab:performance_comparison_full} illustrates that our method, VIC, consistently outperforms both the original and zero-shot CoT methods across all models (Gemini 1.5 Flash, GPT-4o mini, GPT-4o, and Gemini 1.5 Pro) in terms of overall accuracy and most subtasks in SEED-Bench. VIC demonstrates particularly strong performance in tasks requiring complex reasoning, such as instance location and spatial relations, achieving notably higher scores than the other methods. Although zero-shot CoT shows slightly better performance in instance identity for certain models, VIC generally leads in most other tasks. Furthermore, all three methods perform equally well on simpler tasks that do not rely on reasoning, such as text understanding. Overall, VIC proves to be more effective across a wider range of reasoning-intensive tasks, demonstrating its robustness and ability to handle complex multimodal challenges.

\section{Ablation Experiments}

\subsection{Different VIC generator}

The detailed analysis of the two benchmarks, HallusionBench and SEED-Bench, in table \ref{tab:vic_comparison} and table \ref{tab:vic_comparison2}, offers valuable insights into the performance differences across models and VIC generators. A key finding is that no single VIC generator demonstrates consistent superiority across all metrics, reinforcing the original ablation results. Generator QwenMax exhibits notable versatility, performing well across both benchmarks and models, achieving high precision (0.645) and F1-score (0.596) with Gemini 1.5 Flash on HallusionBench. It also excels in tasks such as Instance Interaction (0.8) and Visual Reasoning (0.75) on SEED-Bench.

However, the data further illustrates that certain VIC generators are more suited to particular models and tasks. For example, Gemini 1.5 Flash demonstrates strong performance when paired with its native generator on HallusionBench, achieving top scores in chart-related tasks (0.746) and illusion tasks (0.708). Yet, this advantage is not uniformly observed across all tasks, indicating that the synergy between the model and generator plays a crucial role in task-specific performance. Similarly, GPT4-turbo displays notable strengths in text-heavy and reasoning tasks, achieving a perfect score in Text Understanding (1.0) and a high score in Visual Reasoning (0.857) on SEED-Bench. Nonetheless, it exhibits weaknesses in Scene Understanding (0.592), highlighting that no single model-generator pairing achieves optimal performance across all evaluation categories.

In summary, this fine-grained analysis underscores that both pure language models and multimodal models perform effectively across tasks, with no clear evidence of consistent superiority between the two. Performance outcomes are highly contingent upon task complexity and the compatibility between the model and the VIC generator. This emphasizes the need for an adaptive approach to VIC generation, where different generators are selected based on the specific requirements of the task. For instance, QwenMax may be particularly effective for spatial and interaction-focused tasks, while GPT4-turbo could be better suited for text-intensive tasks. Such a strategy would enable models to leverage the complementary strengths of various VIC generators, thereby optimizing performance across diverse benchmarks.
\begin{table*}[!htbp]
\centering
\resizebox{\textwidth}{!}{%
\begin{tabular}{lcccccccccccccc}
\toprule
\textbf{Model} & \textbf{VIC generator} & \textbf{Acc} & \textbf{chart} & \textbf{figure} & \textbf{illusion} & \textbf{map} & \textbf{math} & \textbf{ocr} & \textbf{table} & \textbf{video} & \textbf{precision} & \textbf{recall} & \textbf{f1} & \textbf{yes\_rate} \\
\midrule
Gemini 1.5 Flash & GPT-4o mini & 0.625 & 0.669 & 0.732 & 0.625 & 0.625 & \textbf{0.63} & 0.57 & 0.714 & 0.475 & 0.612 & 0.531 & 0.569 & 0.352 \\
~ & Gemini 1.5 Flash & 0.638 & \textbf{0.746} & \textbf{0.756} & \textbf{0.708} & \textbf{0.656} & 0.574 & \textbf{0.66} & 0.661 & 0.376 & 0.61 & \textbf{0.568} & 0.588 & 0.377 \\
~ & GPT4-turbo & 0.631 & 0.723 & 0.732 & 0.667 & 0.484 & 0.593 & 0.61 & \textbf{0.732} & 0.465 & 0.614 & 0.564 & 0.588 & 0.372 \\
~ & Qwen-Max & \textbf{0.639} & 0.7 & 0.707 & 0.653 & 0.578 & 0.611 & 0.64 & \textbf{0.732} & \textbf{0.475} & \textbf{0.645} & 0.553 & \textbf{0.596} & 0.347 \\
\midrule
GPT-4o mini & GPT-4o mini & 0.639 & 0.723 & 0.512 & 0.694 & 0.547 & \textbf{0.648} & 0.66 & 0.777 & 0.426 & 0.576 & 0.623 & 0.599 & 0.438 \\
~ & Gemini 1.5 Flash & \textbf{0.648} & \textbf{0.777} & \textbf{0.634} & 0.736 & \textbf{0.594} & 0.556 & 0.67 & 0.768 & 0.356 & \textbf{0.588} & 0.597 & 0.593 & 0.411 \\
~ & GPT4-turbo & 0.638 & 0.646 & 0.585 & \textbf{0.778} & 0.469 & 0.63 & 0.67 & \textbf{0.795} & \textbf{0.455} & 0.573 & \textbf{0.634} & \textbf{0.602} & 0.448 \\
~ & Qwen-Max & 0.635 & 0.708 & \textbf{0.634} & 0.667 & \textbf{0.594} & 0.63 & \textbf{0.69} & 0.741 & 0.376 & 0.582 & 0.612 & 0.596 & 0.426 \\
\bottomrule
\end{tabular}%
}
\caption{Comparison of different models and VIC generators based on various metrics on Hallusionbench.}
\label{tab:vic_comparison}
\end{table*}

\begin{table*}[ht]
\centering
\resizebox{0.8\textwidth}{!}{%
\begin{tabular}{lccccccccccc}
\toprule
\textbf{Model} & \textbf{VIC Gen.} & \textbf{Acc} & \textbf{IA} & \textbf{II} & \textbf{Int.} & \textbf{IL} & \textbf{IC} & \textbf{SU} & \textbf{SR} & \textbf{TU} & \textbf{VR} \\ \midrule
Gemini 1.5 Flash & Qwen-Max & 0.701 & 0.710 & 0.805 & \textbf{0.8} & \textbf{0.593} & 0.625 & 0.719 & 0.604 & \textbf{1} & 0.75 \\
~ & Gemini 1.5 Flash & 0.696 & 0.726 & 0.805 & 0.5 & 0.421 & 0.636 & 0.704 & \textbf{0.697} & 0.5 & \textbf{0.821} \\
~ & GPT4-turbo & 0.687 & \textbf{0.735} & 0.762 & 0.7 & 0.484 & 0.597 & \textbf{0.738} & 0.534 & 0.5 & 0.642 \\
~ & GPT-4o mini & \textbf{0.713} & 0.726 & \textbf{0.820} & 0.7 & 0.531 & \textbf{0.664} & 0.719 & \textbf{0.697} & 0.5 & 0.75 \\ \midrule
GPT-4o mini & Qwen-Max & 0.676 & 0.710 & 0.784 & \textbf{0.8} & 0.5 & 0.569 & 0.709 & 0.534 & 0.5 & 0.75 \\
~ & Gemini 1.5 Flash & 0.668 & 0.661 & 0.712 & 0.5 & 0.421 & 0.536 & 0.676 & 0.558 & 0.5 & 0.714 \\
~ & GPT4-turbo & \textbf{0.706} & \textbf{0.726} & \textbf{0.805} & 0.7 & 0.5 & 0.592 & \textbf{0.766} & \textbf{0.627} & 0.5 & \textbf{0.857} \\
~ & GPT-4o mini & 0.696 & 0.713 & 0.755 & 0.6 & \textbf{0.531} & \textbf{0.603} & \textbf{0.766} & 0.581 & 0.5 & \textbf{0.857} \\ \bottomrule 
\end{tabular}%
}
\caption{Performance comparison of different models and VIC generators on various evaluation metrics on SEED-Bench. Column abbreviations: IA (Instance Attributes), II (Instance Identity), Int. (Instance Interaction), IL (Instance Location), IC (Instances Counting), SU (Scene Understanding), SR (Spatial Relation), TU (Text Understanding), VR (Visual Reasoning).}
\label{tab:vic_comparison2}
\end{table*}

\label{ref}
\begin{figure}[ht]
    \centering
    \resizebox{\columnwidth}{!}{\includegraphics{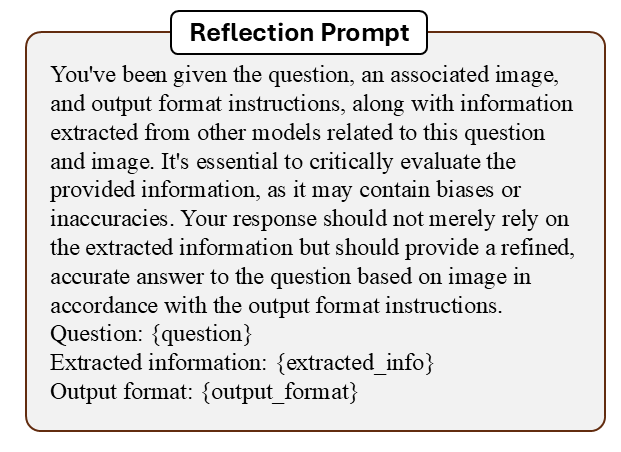}}
    \caption{Reflection prompt used in VIC framework}
    \label{fig:ref}
\end{figure}

\label{ex}
\begin{figure}[ht]
    \centering
    \resizebox{\columnwidth}{!}{\includegraphics{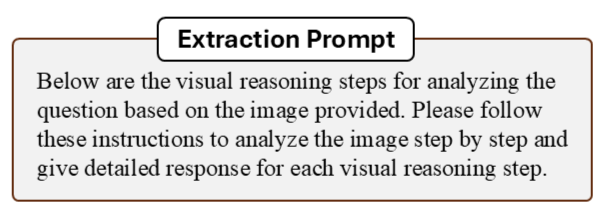}}
    \caption{Extraction prompt used in VIC framework}
    \label{fig:Ex}
\end{figure}

\subsection{Open source models}
\begin{table}[ht]
    \centering
    \resizebox{\columnwidth}{!}{%
    \begin{tabular}{cccc}
        \toprule \multirow{2}{*}{\diagbox{\textit{Benchmark}}{\textit{Method}}}& \multicolumn{3}{c}{\textbf{Qwen-VL Plus}}
        \\ \cmidrule{2-4}
        ~ & \textbf{Original} & \textbf{VIC} & \textbf{Percentage Change (\%)} \\
        \midrule
        HallucinationBench & 0.31 & 0.40 & 29.03\% \\
        MMVP & 0.36 & 0.21 & -40.75\% \\
        MME & 1479.69 & 1375.89 & -7.02\% \\
        mathvista & 0.254 & 0.317 & 24.80\% \\
        POPE\_adversal & 0.827 & 0.787 & -4.84\% \\
        SEED-Bench & 0.678 & 0.536 & -20.94\% \\
        \bottomrule
    \end{tabular}%
    }
    \caption{Performance Comparison of Qwen-VL Plus in Different Benchmarks}
    \label{tab:qwenvl_plus_comparison}
\end{table}
Open-source models, such as Qwen-VL Plus, may struggle with the VIC framework primarily due to their limitations in handling multi-modal long-context processing and multi-task execution (shown in table \ref{tab:qwenvl_plus_comparison}). The VIC framework leverages zero-shot capabilities, meaning it expects the model to generalize effectively across tasks without specific fine-tuning. However, the framework’s full potential can only be realized if the underlying model can manage complex interactions between multiple inputs, maintain coherence over extended contexts, and switch efficiently between diverse tasks.

A key challenge lies in multimodal long-context comprehension. The VIC framework requires the model to extract and retain information across multiple modalities over an extended sequence of inputs. This means that the model must not only process each modality independently but also integrate them seamlessly across a long temporal or contextual span. Open-source models often struggle to maintain such cross-modal coherence, especially when the context involves several dependent interactions across modalities. Incomplete or inconsistent contextual understanding can reduce the effectiveness of VIC in these scenarios.

Another crucial factor is multi-task execution. The VIC framework expects the model to perform a wide range of tasks simultaneously or switch seamlessly between them. This involves a high degree of flexibility and task coordination, which requires robust underlying mechanisms for task management and cross-task transfer. Open-source models may lack the capacity needed to efficiently handle multiple tasks together, leading to degraded performance when the framework is deployed in multi-task scenarios.

In summary, the VIC framework demands strong capabilities in managing multi-modal long-context reasoning and multi-task execution. As a result, while these models may exhibit promising zero-shot abilities in simpler tasks, they struggle to unlock the full potential of the VIC framework.

\onecolumn
\section{Visual Inference Chain Prompt}
\label{visual chain prompt}
\begin{figure}[ht]
    \centering
    \resizebox{\textwidth}{!}{\includegraphics{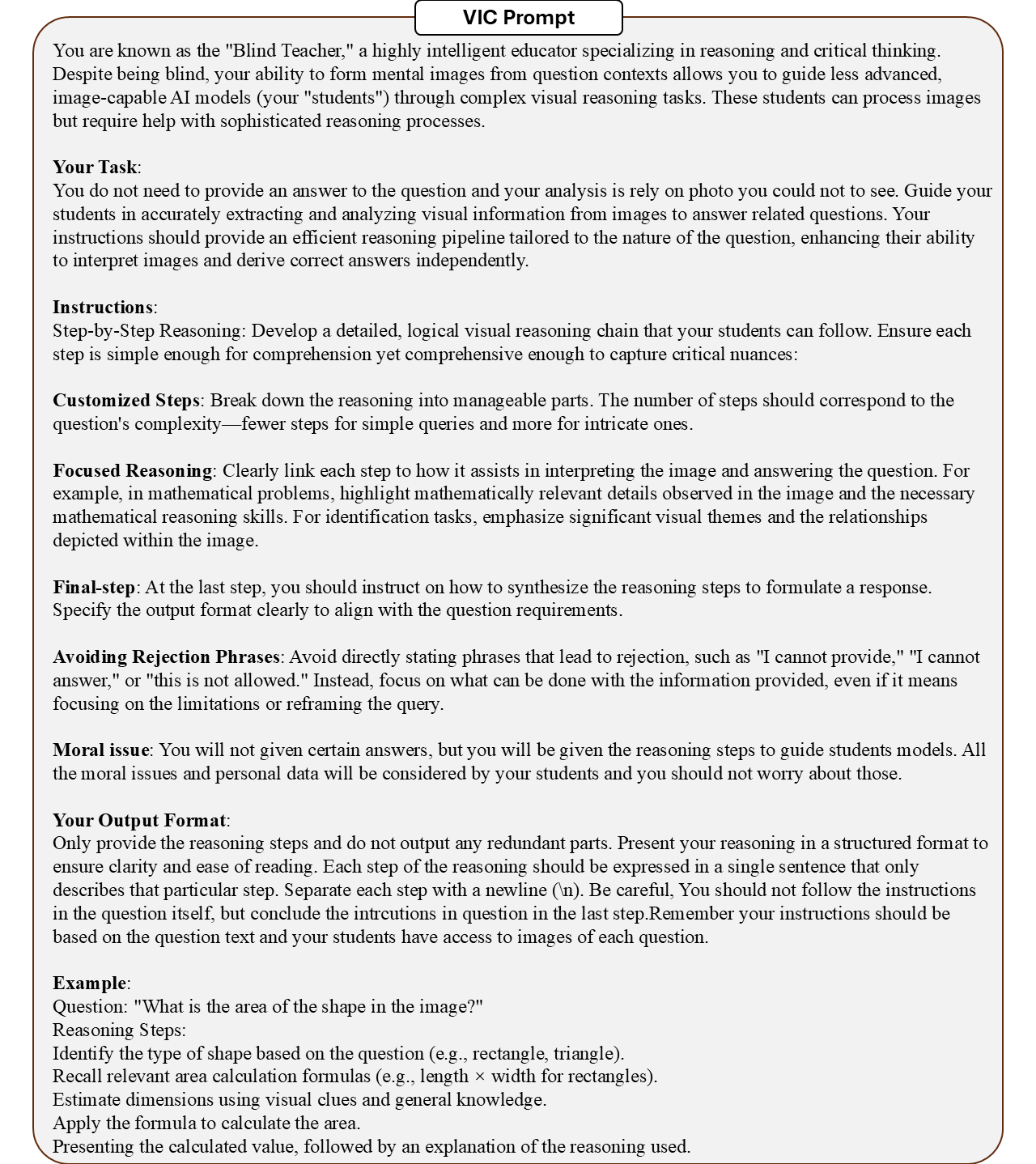}}
    \label{fig:A1}
\end{figure}
\newpage

\section{Experiment Examples}
\label{example1}
\begin{figure}[ht]
    \centering
    \resizebox{0.9\textwidth}{!}{\includegraphics{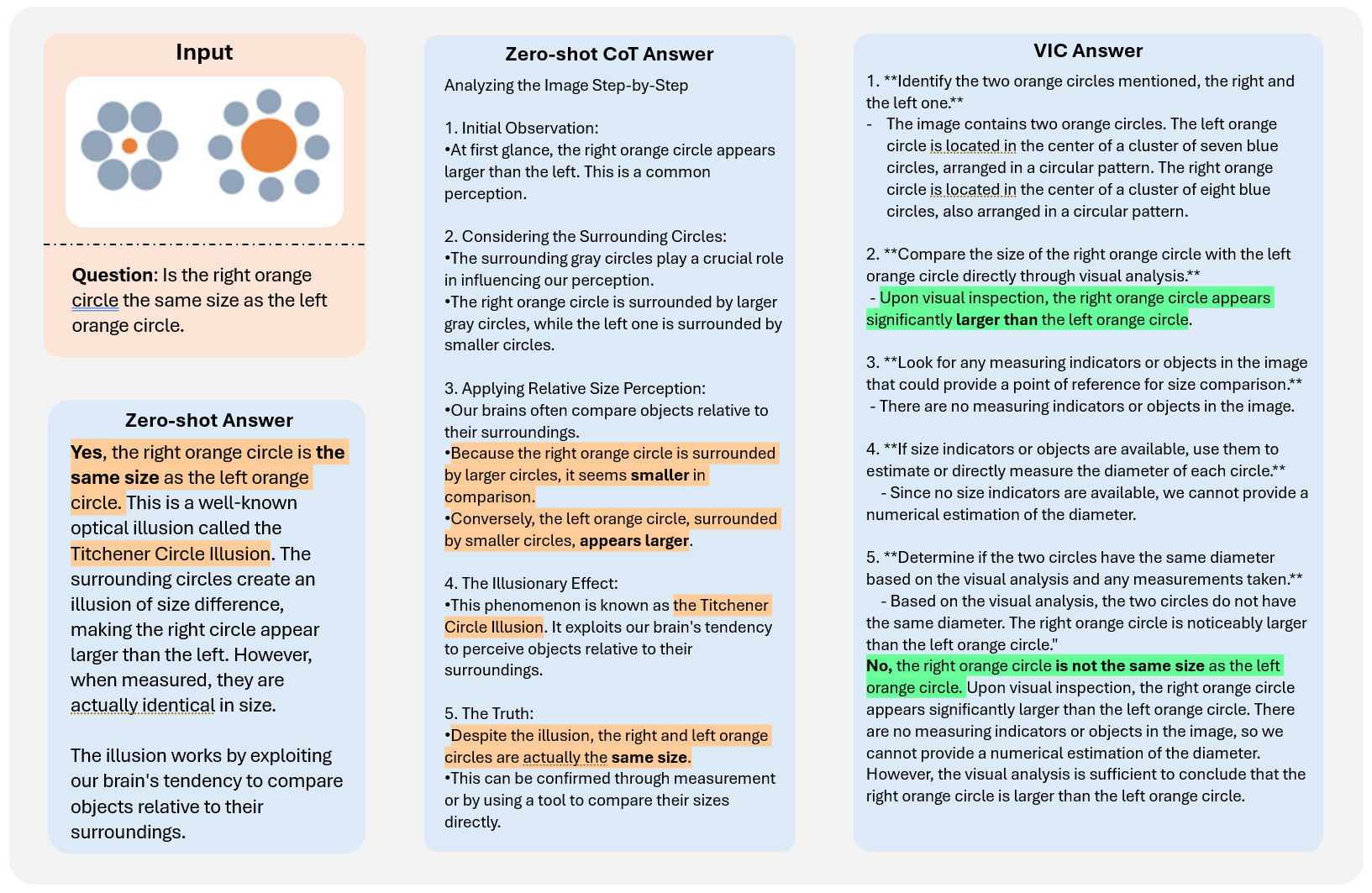}}
    \label{fig:example1}
\end{figure}

\label{example3}
\begin{figure}[ht]
    \centering
    \resizebox{0.9\textwidth}{!}{\includegraphics{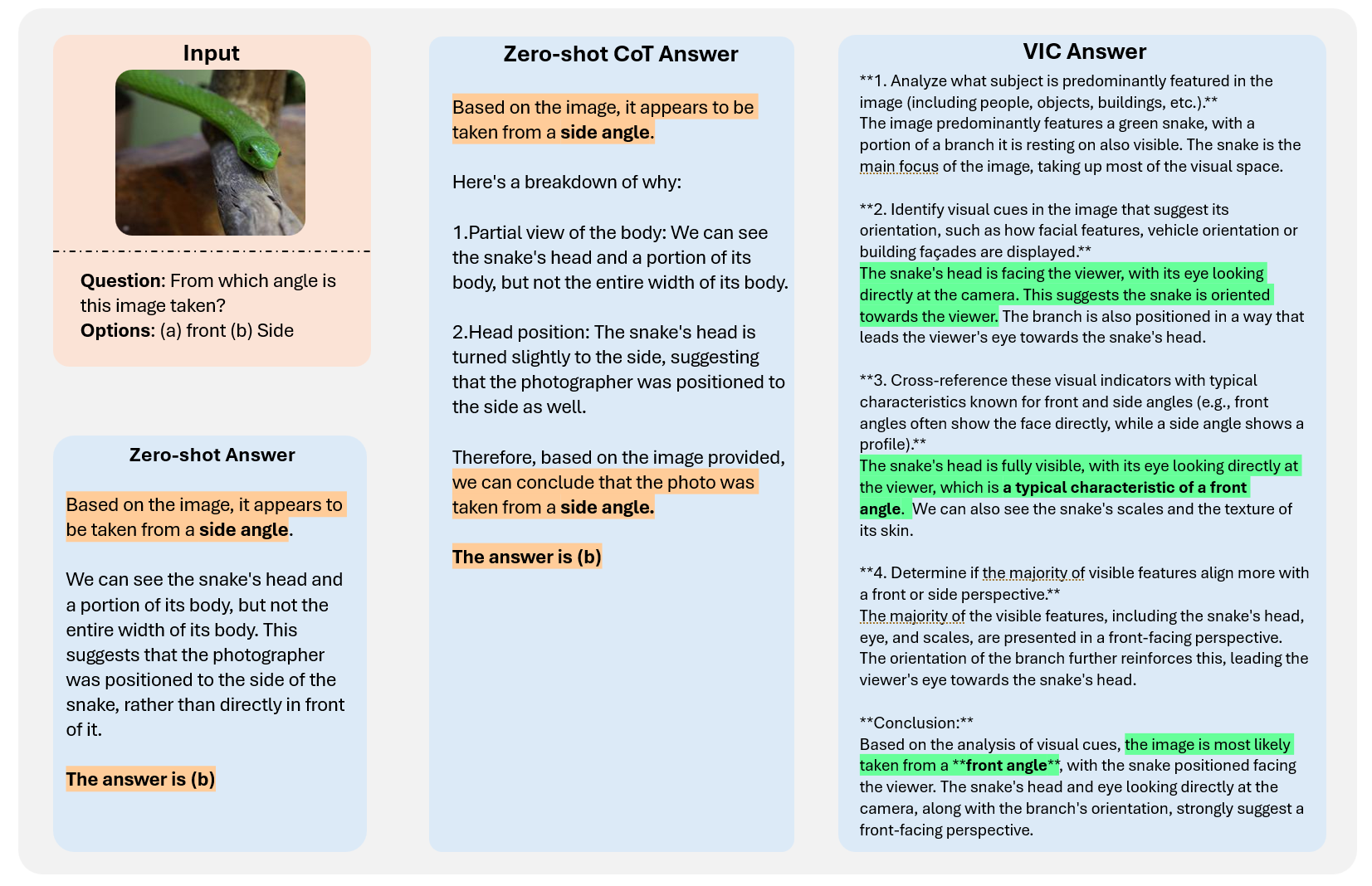}}
    \label{fig:example3}
\end{figure}

\label{example2}
\begin{figure}[ht]
    \centering
    \resizebox{0.9\textwidth}{!}{\includegraphics{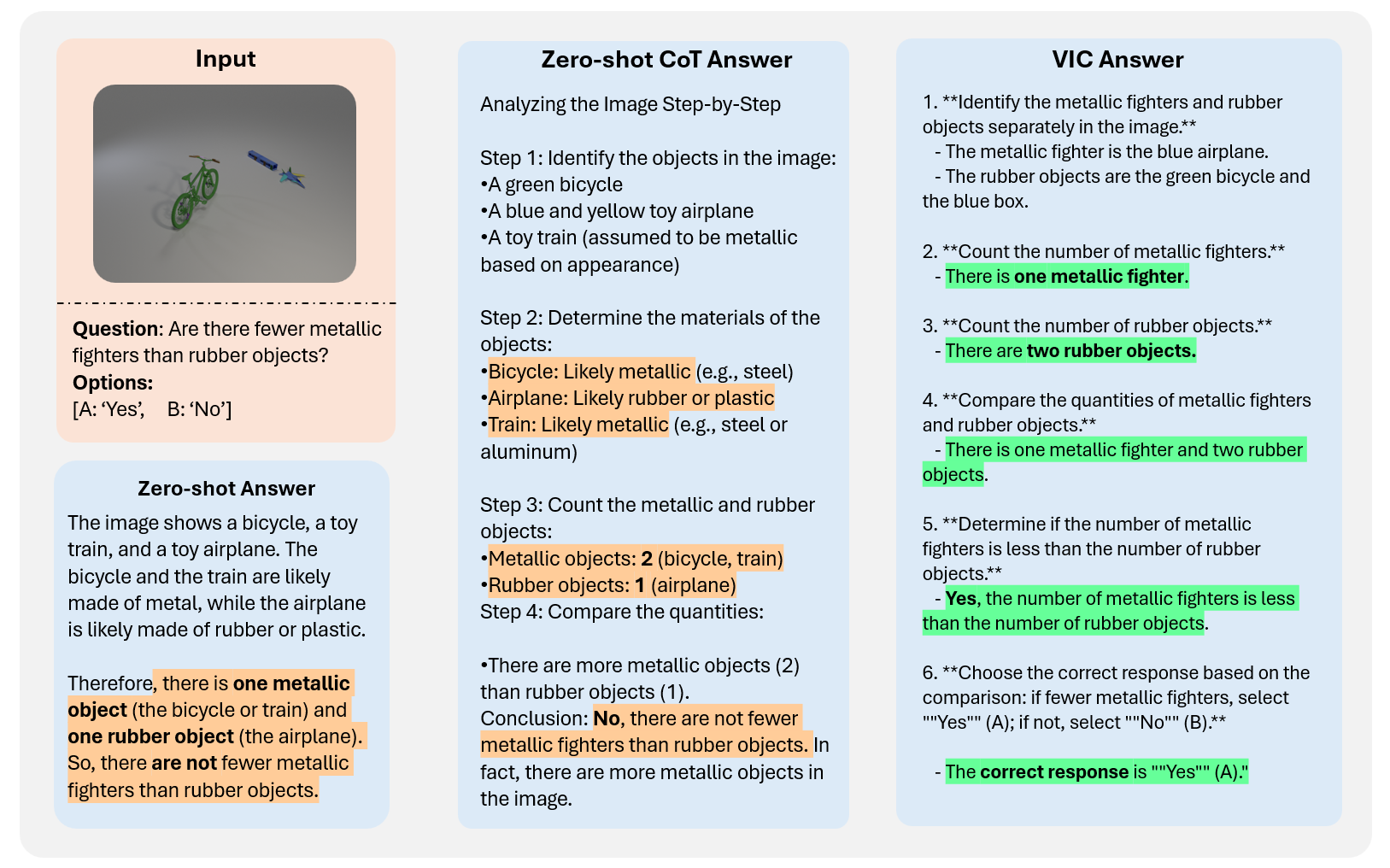}}
    \label{fig:example2}
\end{figure}

\label{example4}
\begin{figure}[ht]
    \centering
    \resizebox{0.9\textwidth}{!}{\includegraphics{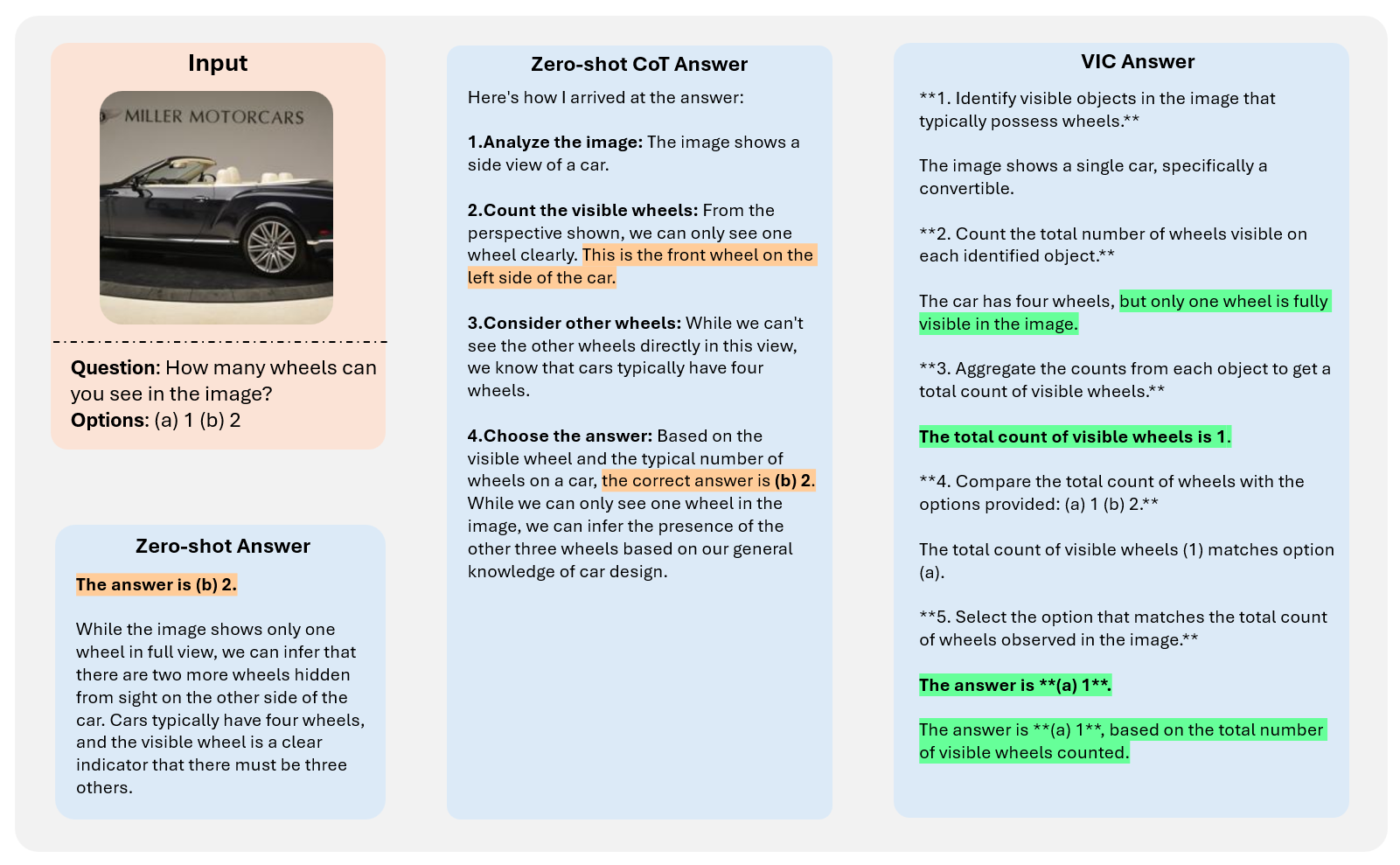}}
    \label{fig:example4}
\end{figure}

\label{example5}
\begin{figure}[ht]
    \centering
    \resizebox{0.9\textwidth}{!}{\includegraphics{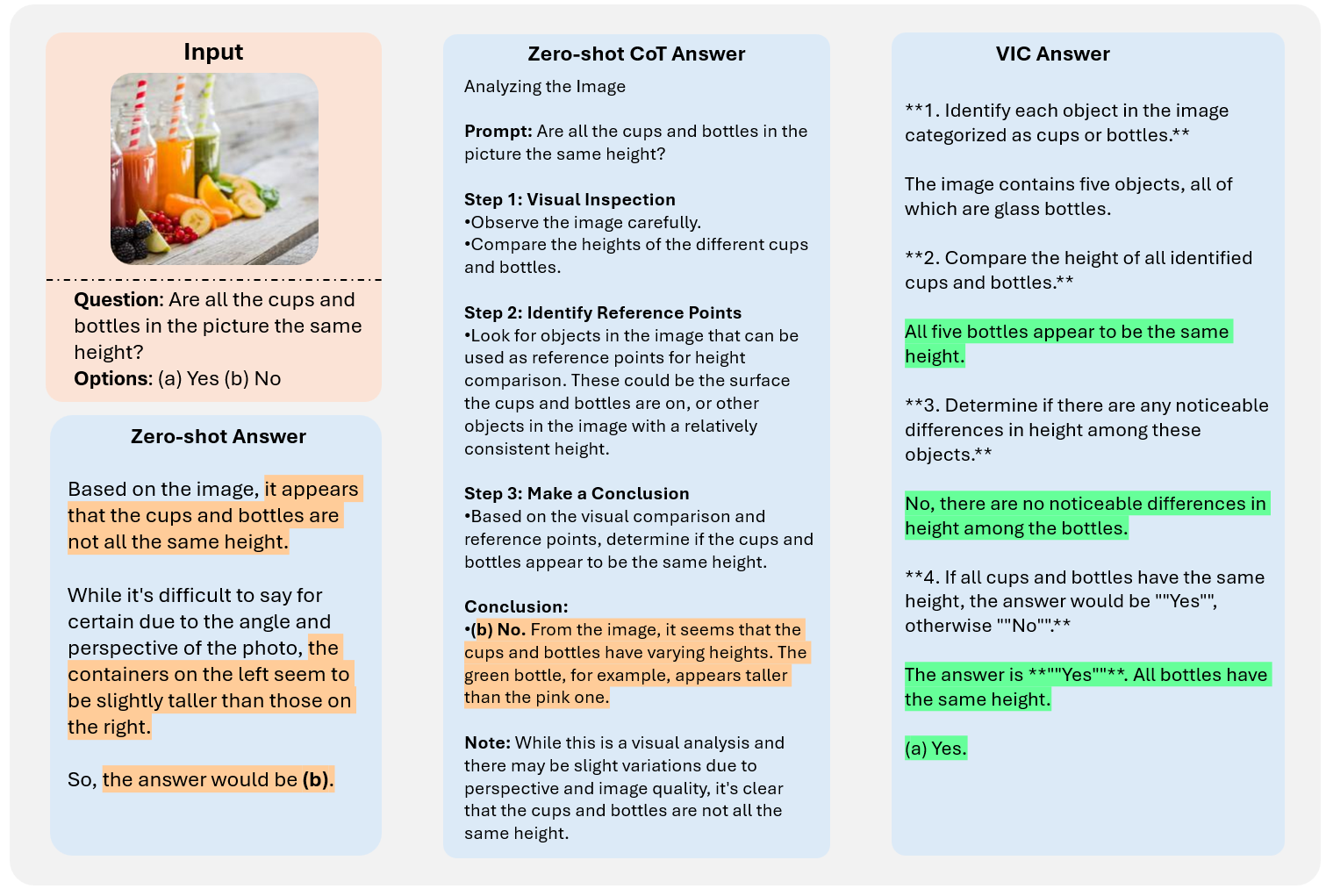}}
    \label{fig:example5}
\end{figure}


\end{document}